\pdfoutput=1
\documentclass{article}
\usepackage{iclr2027_conference}
\usepackage[T1]{fontenc}
\renewcommand{\sfdefault}{phv}

\usepackage{amsmath,amsfonts,bm}

\def\eqref#1{equation~\ref{#1}}

\def\1{\bm{1}}

\DeclareMathAlphabet{\mathsfit}{\encodingdefault}{\sfdefault}{m}{sl}
\SetMathAlphabet{\mathsfit}{bold}{\encodingdefault}{\sfdefault}{bx}{n}

\usepackage{hyperref}
\hypersetup{hidelinks}
\usepackage{url}
\usepackage{graphicx}
\usepackage{booktabs}
\usepackage{array}
\usepackage{microtype}
\usepackage{placeins}
\usepackage{tabularx}
\usepackage{tikz}
\usetikzlibrary{arrows.meta,positioning,fit}
\newcolumntype{Y}{>{\raggedright\arraybackslash}X}
\usepackage{xspace}
\title{OSCC: Certified Observation-Safe Coupling Optimization
for Gradient-Noise Control in Imperfect-Information Learning}
\author{Miaobo Hu$^{1,2,*}$, Shuhao Hu$^{2}$, Xiaobo Guo$^{2}$, Xin Wang$^{2}$,\\
Bokun Wang$^{2}$, Rui Chen$^{2}$, Daren Zha$^{2}$\\[4pt]
{\normalfont $^{1}$School of Artificial Intelligence, University of Chinese Academy of Sciences, Beijing, China}\\
{\normalfont $^{2}$Institute of Information Engineering, Chinese Academy of Sciences, Beijing, China}\\
{\normalfont $^{*}$Corresponding author: \texttt{humiaobo@iie.ac.cn}}}

\iclrfinalcopy
\hypersetup{pdfauthor={Miaobo Hu, Shuhao Hu, Xiaobo Guo, Xin Wang, Bokun Wang, Rui Chen, Daren Zha}}
\begin{document}
\maketitle
\lhead{Preprint}

\begin{abstract}
Coupled rollouts can reduce the noise of counterfactual action comparisons, but
two issues prevent standard common-random-number constructions from serving as
a general learning primitive in imperfect-information environments. First, an
invalid coupling may expose hidden state, synchronize endogenous policy
randomness, or misalign chance events after counterfactual histories diverge.
Second, in multi-action policy optimization, lower return-contrast variance is
not by itself the relevant objective: the optimizer depends on the return
covariance matrix after projection through the local policy-gradient geometry.

We introduce observation-safe counterfactual coupling (OSCC), a framework that
defines an admissible class through marginal preservation, information-state
safety, branch-local policy randomness, semantic event alignment, and
trace-before-oracle replay. We derive a gradient-aware coupling criterion
showing that, for marginal-preserving couplings, policy-gradient noise changes
are determined by policy-Jacobian-weighted off-diagonal return covariance. This
motivates OSCC-Select, a calibration-only selector that chooses among
independent, root-only, continuation-only, and fully coupled rollouts using
separate safety and gain certificates. Its gain target combines projected
gradient noise with measured physical sampling cost and falls back to
independent sampling whenever a simultaneous lower confidence bound does not
certify improvement.

The fully coupled CP-GRPO construction is one OSCC instantiation. On 100,000
fixed-root Leduc comparisons, it reduces return-contrast variance from 41.1158
to 18.1441, a 55.87\% reduction, while preserving the declared branch
marginals. The completed policy, solver, transfer, selector, safety,
noise-transfer, and physical-cost surfaces separate coupling validity,
estimator precision, and the downstream optimizer response. With three
actions, OSCC-Select chooses continuation coupling and attains gradient-noise
trace 0.0783 versus 0.0917 for return-variance selection. Increasing calibration
from 64 to 2,048 groups raises certification from 0.327 to 0.995.

\end{abstract}

\section{Introduction}
Rollout comparisons in imperfect-information games spend variance on chance
events that are unrelated to the root-action contrast of interest. A fixed
root deal and a common continuation stream can cancel part of this nuisance
variation, but a shared random source can also expose hidden state, synchronize
endogenous policy draws, or reuse a chance event after two histories diverge.
Even when every branch marginal is preserved, the coupling that minimizes a
scalar return contrast need not minimize the noise of a multi-action policy
update. The relevant object is the return covariance matrix after projection
through the local policy-gradient geometry.

We formulate OSCC as a constrained coupling problem. The framework defines the
admissible class under partial observation, then uses OSCC-Select to estimate an
update-relevant noise functional and physical sampling cost on a calibration
split. Candidate gains are certified with simultaneous confidence bounds; the
mapping is frozen before held-out learning and falls back to independent
sampling when no improvement is certified. CP-GRPO is the fully
coupled instantiation used to expose the covariance mechanism. Its rollout
construction groups legal root actions, derives semantic continuation streams,
and keeps policy-side action randomness branch-local.

The fixed-root audit measures a 55.87\% reduction in return-contrast variance
over 100,000 Leduc comparisons. Root-only and continuation-only controls
separate the two covariance sources, while the three-seed diagnostic
has lower final diagnostic exploitability for independent sampling on every
seed. The observation audit, branch-swap replay, clipping screen, and the
completed selector, cost, solver, and transfer surfaces keep estimator and
learning claims on separate statistical units.

Our contributions are:
\begin{enumerate}
\item \textbf{Observation-safe coupling class.} We formalize admissible
counterfactual couplings through marginal preservation, information-state
safety, branch-local policy randomness, semantic event alignment, and
trace-before-oracle ordering.
\item \textbf{Gradient-aware coupling characterization.} We show that,
among marginal-preserving couplings, multi-action policy-gradient noise is
changed by policy-gradient-weighted off-diagonal return covariance; the
binary-action case reduces to return-contrast variance.
\item \textbf{Certified adaptive selection.} OSCC-Select chooses among
independent, root-only, continuation-only, and fully coupled rollouts using
separate safety and simultaneous cost-adjusted gain certificates, with an
independent fallback.
\item \textbf{Estimator-to-optimizer evaluation.} We specify the complete
chain from marginal validity and covariance to projected gradient noise,
realized update variance, physical cost, held-out quality, solver evaluation,
and cross-environment transfer, with explicit statistical units.
\end{enumerate}

\section{Related Work}
Counterfactual regret minimization and its neural variants use
counterfactual reach probabilities and regret updates in imperfect-information
games \citep{zinkevich2007regret,tammelin2015solving,brown2019deepcfr}.
NFSP, OpenSpiel, ReBeL, DeepStack, Libratus, and Pluribus establish
reproducible self-play or search interfaces at different scales
\citep{heinrich2016nfsp,lanctot2019openspiel,brown2020rebel,
moravcik2017deepstack,brown2018libratus,brown2020pluribus}. These systems motivate exploitability-oriented evaluation. Our comparison
unit is a legal root-action group with explicitly shared exogenous draws.

The closest statistical precedents are explicit and substantive. PEGASUS
uses common scenarios for policy search, and the paired-comparison framework
of Strens and Moore studies how shared scenarios reduce comparison noise
\citep{ng2000pegasus,strens2002paired}. The vine procedure in TRPO also
evaluates alternative actions from a common state and reports the resulting
value differences \citep{schulman2015trpo}. More recent rollout work applies
common random numbers to simulation-based planning
\citep{yadav2026crnrollouts}. These papers establish common scenarios as a useful comparison device.
The additional constraint in our setting is that a shared full root contains
hidden cards: both the observation projection and the policy random-stream
address must preserve the acting player's information state.

PPO and TRPO stabilize policy updates through clipped or trust-region
objectives \citep{schulman2017ppo,schulman2015trpo}, while group-relative
objectives normalize sampled return groups \citep{shao2024deepseekmath,
qwenmath2024}. OSCC is complementary: it changes how trajectories are coupled
before a contrast is formed and leaves the action space and policy
parameterization unchanged. CP-GRPO is the fully coupled GRPO-style
instantiation used in the fixed-root audit. Table~\ref{tab:prior-art}
compares the objectives and information contracts of these methods.

AIVAT uses control variates and known strategies to reduce evaluation
variance in imperfect-information games \citep{burch2018aivat}. CP-GRPO
instead couples action alternatives before their return difference is formed.

The statistical identity is classical: sharing exogenous draws can reduce
the variance of a difference when the induced covariance is positive without
changing either marginal target \citep{glasserman2004monte}. Our contribution
is the operational contract around that identity: root, continuation, and
policy keys; observation-only inputs; early termination; seat-swapped
duplicates; and a deterministic evidence ledger. The implementation makes an
estimator claim separable from a learning claim, and the ablation
measures which part of the coupling is responsible for the observed covariance.

\paragraph{What OSCC adds to common-random-number pairing.}
Classical paired simulation asks whether shared exogenous scenarios reduce
comparison noise. OSCC additionally specifies which events may be shared after
counterfactual histories diverge, keeps endogenous policy draws branch-local,
and records the observation and replay predicates needed to audit the joint
trajectory law. The selector therefore distinguishes an admissible but harmful
coupling from an invalid coupling that only appears beneficial. These are
policy-facing constraints around the classical covariance identity rather than
a new variance identity.

\section{Method}
\subsection{Task definition}
We consider a finite two-player imperfect-information game. A simulator state
$s$ contains public history, private cards, the acting seat $p$, and legal
actions $\mathcal A(I)$; the policy sees the projection $I=I(s,p)$.
A comparison group fixes a root distribution and evaluates each legal action
under a fixed continuation policy. Its target is the action-value difference
$Q(I,a)-Q(I,b)$, where the expectation integrates hidden deals compatible with
$I$ and subsequent chance events. Training roots, validation roots, and test
roots have separate stream namespaces.

The collector returns branch utilities, legal-action masks, behavior
probabilities, and sampling counts. A terminal branch contributes its terminal
utility and consumed cost. A malformed action is retained with a failure code.
Solver outputs enter the evaluation record after the action trace is frozen.
The policy parameters are learned; the serializer, legal-action rules,
continuation budget, and evaluation solver are fixed within a comparison.

\subsection{Observation-safe coupling}
Let an action group $g$ contain legal root actions
$\{a_{g,i}\}_{i\in I_g}$. All branches begin from the same root state. The
private deal and root seat are shared; after the intervention, chance draws
and policy action draws use distinct counter streams. We make the dependency
explicit:
\[
 k^{\rm env}_{g,e}=H(\texttt{run},\texttt{root},g,e),\qquad
 k^{\rm pol}_{g,b,t}=H(\texttt{run},g,b,t,\texttt{policy}),
\]
where $k^{\rm env}$ may regenerate the hidden simulator state, while
$k^{\rm pol}$ contains no private-card or full-state field. The policy
lookup key is
\[
k_t=I(s_t,p_t)=\texttt{P}p_t:\texttt{private}(p_t):
\texttt{public}:\texttt{round}:\texttt{history}:\texttt{legal-mask}.
\]
The policy lookup $k_t$ excludes the opponent card, terminal oracle, and
branch identity. The sampler uses branch identity only as an opaque address
for independent policy draws; its address excludes hidden-state fields. A seat-swapped duplicate
is generated from an explicitly copied root state and is recorded as a
separate group.

Table~\ref{tab:rng-dependency} summarizes the dependency contract. The
distinction between a coupling key and a physical sampling call matters:
one key addresses a deterministic stream, whereas a sampler can consume
several draws before the next event. We therefore report both quantities in
the completed ablation and never describe a shared key count as a call count.

\begin{table}[!htbp]
\centering
\scriptsize
\setlength{\tabcolsep}{3pt}
\caption{RNG dependency contract. Shared environment randomness is allowed
only through legal observations; policy-side addressing is branch-specific
and excludes hidden state.}
\label{tab:rng-dependency}
\begin{tabularx}{\columnwidth}{@{}Y Y Y Y@{}}
\toprule
namespace & inputs & shared? & policy visibility and audit \\
\midrule
replay/root & run, root, group & yes & no; root hash equality \\
environment & env key, event counter & yes & legal observation only; mask/hash check \\
continuation & env key, public history & yes & through public events; monotone counter \\
policy action & run, group, branch, time & no & action draw only; hidden-card invariance \\
ledger & event, branch, status & no & no effect on rollout; join after trace freeze \\
\bottomrule
\end{tabularx}
\end{table}

The implementation records the logical group key, root-state hash,
observation hash, continuation-stream key, and policy-stream key. A replay
regenerates the same event order and verifies that swapping branch
identifiers leaves the policy-visible information state unchanged. Changing
only $k^{\rm pol}$ may change the sampled action; the observation,
legal-action mask, and replay metadata remain unchanged. The 774-state audit in
the observation audit exercises this invariant and finds no opponent-card
exposure. A branch-swap test checks that the measured contrast changes sign
under a utility-preserving seat swap.

\subsection{Admissible observation-safe couplings}
\label{sec:oscc}
Let $I$ denote the acting player's information state and let
$\mathcal{A}(I)=\{a_1,\ldots,a_m\}$ denote its legal actions. For branch $i$,
write the return as
$R_i=f_i(D,U_i,W_i)$, where $D$ contains hidden root variables compatible
with $I$, $U_i$ contains exogenous continuation randomness, and $W_i$ contains
policy-side action randomness. A coupling $\Gamma$ is a joint distribution over
all branch variables. We call $\Gamma$ \emph{admissible} when it satisfies the
following conditions.

\paragraph{C1: Marginal preservation.}
For every legal intervention $a_i$, the branch marginal under $\Gamma$ equals
the declared simulator-policy marginal $P(D,U,W\mid I,a_i)$. Coupling can
change cross-branch dependence while leaving each branch target unchanged.

\paragraph{C2: Information-state safety.}
At every policy decision time, the online input is an information-state
projection $I_t=\mathcal{I}(s_t,p_t)$. Opponent private state,
counterfactual branch records, terminal oracle labels, and replay-only
identifiers are excluded from the policy-visible filtration.

\paragraph{C3: Policy-randomness separation.}
Environment randomness may be shared, but policy action draws use branch-local
namespaces
\[
k^{\mathrm{pol}}_{g,b,t}=H(\mathrm{run},g,b,t,\mathrm{policy}),
\]
whose address contains no hidden simulator field. Sharing an environment event
therefore does not force the same endogenous action sample in two branches.

\paragraph{C4--C5: Semantic event alignment and trace order.}
A shared chance draw is addressed by its semantic event rather than by an
untyped position in a global stream. Reserved addresses remain reserved after
early termination, and the ordered rollout trace is frozen before solver or
oracle fields are joined.

\paragraph{Semantic-event coupling operator.}
For a compatible event signature $e$ (public history, chance type, round, and
occurrence index), the collector draws one $U_{g,e}\sim\mathrm{Uniform}(0,1)$
and applies each branch's own conditional transition sampler to that uniform.
An event absent from one branch consumes no transition but keeps its reserved
address. Policy actions use independent $V_{g,b,t}$ streams whose addresses
contain the run, group, branch, and action counter but no hidden simulator
field. This construction makes the shareable exogenous event explicit while
preserving branch-local endogenous randomness.

\subsection{Target preservation and coupling variance}
\label{sec:oscc-theory}
\paragraph{Proposition 1 (target preservation).}
For any admissible coupling and legal actions $a_i,a_j$,
\[
\mathbb{E}_{\Gamma}[R_i-R_j\mid I]
=Q(I,a_i)-Q(I,a_j).
\]
Indeed, admissibility preserves the marginal trajectory distribution of every
branch; the coupling changes only their joint dependence.

\paragraph{Proposition 2 (conditional covariance decomposition).}
Let $D$ denote hidden root variables compatible with $I$ and
$m_i(D)=\mathbb{E}[R_i\mid D,I]$. Then
\[
\operatorname{Cov}(R_i,R_j\mid I)
=\operatorname{Cov}(m_i(D),m_j(D)\mid I)
+\mathbb{E}\left[\operatorname{Cov}(R_i,R_j\mid D,I)\mid I\right].
\]
The first term is induced primarily by root reuse and the second by shared
continuation randomness. Their interaction need not be additive when an
intervention changes the semantic chance events reached by a branch.

For an admissible coupling with coupling-invariant marginal variances,
\[
\operatorname{Var}_{\Gamma}(R_i-R_j\mid I)
=\sigma_i^2(I)+\sigma_j^2(I)-2\operatorname{Cov}_{\Gamma}(R_i,R_j\mid I).
\]
Relative to independent sampling, positive covariance is therefore a measured
variance gain, while a non-positive covariance is a reason to select a weaker
coupling or the independent fallback.

\subsection{From return covariance to gradient noise}
\label{sec:gradient-covariance}
For a legal-action group, collect branch returns in
$\mathbf{R}=[R_1,\ldots,R_m]^\top$ with conditional covariance
$\Sigma_R^\Gamma(I)$. Define
\[
\mathbf{G}_\theta(I)=\left[
\nabla_\theta\pi_\theta(a_1\mid I),\ldots,
\nabla_\theta\pi_\theta(a_m\mid I)
\right].
\]
For the complete-action, unclipped estimator with frozen branch returns,
\[
\Sigma_{\nabla J}^{\Gamma}(I)
=\mathbf{G}_\theta(I)\Sigma_R^\Gamma(I)\mathbf{G}_\theta(I)^\top,
\qquad
\mathcal{N}(\Gamma;I)=\operatorname{tr}\!\left(
\Sigma_{\nabla J}^{\Gamma}(I)\right).
\]
For two actions this reduces to the contrast-variance term. With more than two
actions, the complete off-diagonal covariance matrix matters. Clipping changes
the effective coefficients, so the clipping-support audit is reported as a
separate measurement.

\paragraph{Proposition 3 (gradient-aware coupling difference).}
\label{prop:geometry}
Let $g_i=\nabla_\theta\pi_\theta(a_i\mid I)$. For two admissible couplings
$\Gamma,\Gamma'$ with identical branch marginals and a fixed checkpoint,
\begin{equation}
\mathcal N(\Gamma;I)-\mathcal N(\Gamma';I)
=2\sum_{i<j}\langle g_i,g_j\rangle
\left[\operatorname{Cov}_{\Gamma}(R_i,R_j\mid I)
-\operatorname{Cov}_{\Gamma'}(R_i,R_j\mid I)\right].
\label{eq:geometry-difference}
\end{equation}
Expanding the trace cancels the diagonal terms by marginal preservation.
Consequently, each covariance change is weighted by the local gradient
geometry. For two actions, $g_2=-g_1$, giving
$\mathcal N(\Gamma;I)=\|g_1\|_2^2\operatorname{Var}_{\Gamma}(R_1-R_2\mid I)$.
This exact ranking equivalence holds at fixed $I$ and equal cost; aggregation
across roots must retain the $\|g_1(I)\|_2^2$ weights. With three or more actions,
a scalar contrast can rank candidates differently from $\mathcal N$.
Appendix~\ref{app:selector-proof} supplies the proof and a bounded example.

\subsection{Certified gradient-aware coupling selection}
\label{sec:coupling-selector}
\begin{figure}[!htbp]
\centering
\includegraphics[width=\linewidth]{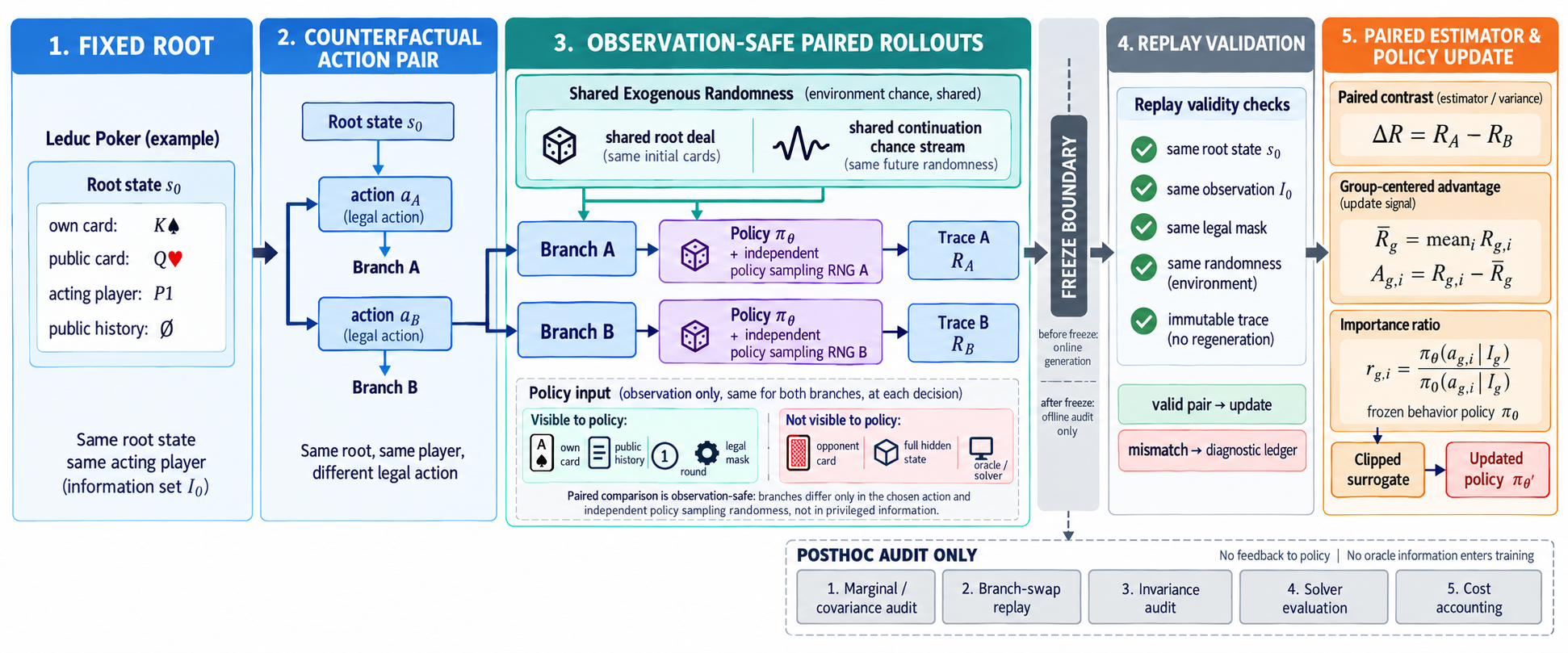}
\caption{CP-GRPO, the full-coupling OSCC instantiation. Blue panels construct
legal action alternatives from one root; the teal panel shares exogenous
chance draws while purple policy streams remain independent. The grey
boundary freezes traces before replay validation and orange policy updates.
Both branches share the root information state; after histories diverge,
each policy receives its own legal information-state projection and chance
events are aligned by semantic address. The lower audit path supplies no
oracle feedback to the online policy.}
\label{fig:method}
\end{figure}

A fixed full coupling is not guaranteed to reduce variance in every
information state. We therefore consider
\[
\mathcal{G}=\{\Gamma_{\mathrm{ind}},\Gamma_{\mathrm{root}},
\Gamma_{\mathrm{cont}},\Gamma_{\mathrm{full}}\}.
\]
For a predeclared stratum $z$ (own-card class, public round, action count, and
acting seat), calibration uses a frozen checkpoint and the declared root law.
The safety gate $\operatorname{Cert}_{\rm safe}(\Gamma,z)$ requires C1--C5.
Define the objective and gain by
\begin{equation}
J_\lambda(\Gamma,z)=\mathbb E_{I\sim z}[\mathcal N(\Gamma;I)]+\lambda C(\Gamma,z),
\qquad \Delta_\lambda(\Gamma,z)=J_\lambda(\Gamma_{\rm ind},z)-J_\lambda(\Gamma,z),
\label{eq:selector-objective}
\end{equation}
where $C$ is expected physical sampler calls per comparison and $\lambda\geq0$
converts calls to gradient-noise units. Cost units and $\lambda$ are fixed before
calibration. A simultaneous lower bound $L_{\Gamma,z}$ defines
$\mathcal E(z)=\{\Gamma:\operatorname{Cert}_{\rm safe}=1,\ L_{\Gamma,z}>0\}$.
OSCC-Select minimizes $\widehat J_\lambda$ over $\mathcal E(z)$, breaking ties by
a fixed arm order, and uses $\Gamma_{\rm ind}$ if $\mathcal E(z)$ is empty.
The selected mapping is frozen before held-out returns or solver values are
observed. Figure~\ref{fig:selector-workflow} details both gates;
Figure~\ref{fig:method} depicts the full-coupling rollout.

\paragraph{Theorem 1 (finite-sample safe fallback).}
\label{thm:fallback}
Suppose C1--C5 hold for eligible candidates, and each of the $M$ declared
candidate--stratum comparisons has $n$ independent calibration blocks with
unbiased gain scores $X_k\in[-B,B]$. Set
\begin{equation}
r_n=B\sqrt{\frac{2\log(2M/\delta)}{n}},\qquad
L_{\Gamma,z}=\overline X_{\Gamma,z}-r_n.
\label{eq:finite-certificate}
\end{equation}
With probability at least $1-\delta$, every selected non-independent arm has
$J_\lambda(\Gamma^\star,z)<J_\lambda(\Gamma_{\rm ind},z)$; fallback gives equality.
A candidate with true gain $d>2r_n$ is certified on the same event. Thus
$n>8B^2\log(2M/\delta)/d^2$ is sufficient. The bound follows from Hoeffding's
inequality and a union bound \citep{hoeffding1963probability}; the score
construction and proof are in Appendix~\ref{app:selector-proof}. The guarantee
concerns the frozen checkpoint and calibration law. Checkpoint, root-law, or
event-interface changes require recertification; unrecognized strata use the
independent arm. Empirical safety checks accompany, rather than replace, the
structural C1--C5 assumptions.

\subsection{CP-GRPO as a Full-Coupling Instantiation}
\label{sec:cpgrpo-instantiation}
CP-GRPO is the full-coupling member of the OSCC family, $\Gamma_{\mathrm{CP}}
=\Gamma_{\mathrm{full}}$. It is not a new policy-optimization objective: the
policy parameterization, action space, and clipped optimization rule remain
unchanged, while the intervention occurs in rollout construction. It reuses a
compatible hidden root and semantic continuation stream while keeping
policy-side draws branch-local. OSCC-Select decides whether this full coupling
should be used at a given checkpoint and stratum; CP-GRPO is the controlled
mechanism probe used to expose the covariance effect.
Relative to standard GRPO, CP-GRPO enumerates the legal root-action group,
freezes branch returns before the update, and applies the complete-action
behavior weights in Eq.~\ref{eq:paired-update}; standard GRPO remains a
sampled-trajectory baseline with its ordinary group construction. The name
therefore identifies the full-coupling instantiation rather than a claim that
the two objectives are identical.
For branch return $R_{g,i}$, the contrast estimator is
$\widehat{\Delta}_g=R_{g,i}-R_{g,j}$. With independent streams,
$\mathrm{Var}(\widehat{\Delta})$ is the sum of marginal variances; with
shared exogenous streams it contains the covariance term
$-2\mathrm{Cov}(R_i,R_j)$. The implementation reports both marginal return
statistics and the paired covariance, so the variance result is not inferred
from a single standard deviation.

The estimator enters a group-centred update after trace freeze. Define
$\bar R_g=|I_g|^{-1}\sum_i R_{g,i}$ and $A_{g,i}=R_{g,i}-\bar R_g$.
For a frozen behavior policy $\pi_0$, let
$r_{g,i}(\theta)=\pi_\theta(a_{g,i}\mid I_g)/\pi_0(a_{g,i}\mid I_g)$.
With complete legal-action enumeration, the clipped surrogate uses behavior
weights:
\begin{equation}
L_g(\theta)=\sum_{i\in I_g}\pi_0(a_{g,i}\mid I_g)
\min\!\left(r_{g,i}A_{g,i},
\operatorname{clip}(r_{g,i},1-\epsilon,1+\epsilon)A_{g,i}\right).
\label{eq:paired-update}
\end{equation}
The returns and centring term are fixed during optimization. For sampled
actions, the collector records the proposal probability and applies the
corresponding sampling weight. This makes the forced-action comparison and
the sampled-action PPO/GRPO controls explicit. Every update stores its ratio,
selected clipping branch, valid transitions, and group denominator.
Appendix~\ref{app:update-details} gives the algebra and support conditions.

\subsection{Training and replay interface}
A run is identified by a SHA-256 manifest. The
online policy receives only public history, its private card, legal actions,
and an independent action draw. Exact solver values and terminal labels are
post-hoc fields. Failed or truncated trajectories remain in the ledger with
an explicit reason and are never silently dropped. The released fixture replays the contract and the summary statistics; the
100,000-group audit is distributed as hash-bound sufficient statistics rather
than a full event log. The manuscript therefore reports the quantities that
are recomputed from the bundle and identifies the aggregate representation in
the artifact record.

A logical coupling key addresses one deterministic stream, while a physical
sampler may consume multiple environment and policy draws. The ledger keeps
both fields, together with branch, event, and terminal status. This prevents
the common error of reporting the number of shared keys as the number of
sampling calls, and lets the completed 2$\times$2 ablation match actual
simulator work rather than only logical group counts.

\section{Experiments}
\subsection{Setup and Research Questions}
The main audit uses fixed-root Leduc groups with 100,000 paired comparisons,
a shared root deal and continuation stream for CP-GRPO, and independent
streams for the control. The authoritative tabular screen uses 40 iterations,
128 sampled decision points per iteration, at most three legal root actions,
two raises per round, and seeds 13, 17, and 23; it retains $5{,}120$ logical
groups per seed. The completed bounded screens use their declared frozen
opponents, root strata, and five-seed scalar connectivity manifest. The
experiments separate five questions:
(Q1) do candidate couplings preserve branch marginals and the observation/replay
contract; (Q2) how do root reuse and continuation coupling change return
covariance and contrast variance; (Q3) does covariance reduction survive
policy-gradient projection, clipping, and the realized update; (Q4) can
OSCC-Select distinguish positive, neutral, adverse, and invalid regimes as
calibration size and physical cost change; and (Q5) do matched policy, solver,
and cross-environment endpoints improve decision quality under the same
budgets? The 100,000-group audit answers Q1--Q2, the clipping and noise-transfer
surfaces answer Q3, and the populated selector stress, policy, solver, and
transfer tables answer the corresponding downstream questions. The multi-action and calibration-size experiments answer Q4 across
binary and ternary action sets and calibration sizes from 64 to 2,048 groups.

\subsection{Main results}
\subsubsection{Return-contrast precision}
\begin{figure}[!htbp]
\centering
\includegraphics[width=0.96\columnwidth]{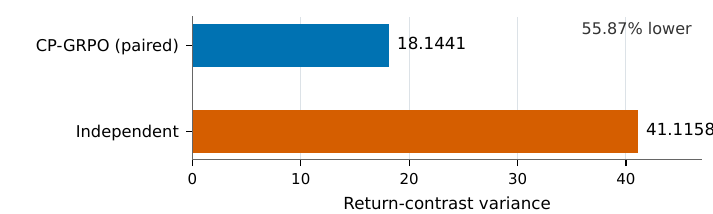}
\caption{Return-contrast variance over 100,000 Leduc comparisons per arm.
Shared root and continuation streams yield 18.1441 versus 41.1158 for
independent draws, a 55.87\% reduction; the aggregate view has no error bars.}
\label{fig:diagnostics}
\end{figure}

Table~\ref{tab:main} reports the primary result. The shared arm has the same
marginal root distribution as the independent arm, but its covariance is
positive for the fixed-root contrast. The variance comparison in Figure~\ref{fig:diagnostics} displays the same
100,000-group result and its relative reduction.

\begin{table}[!htbp]
\centering\scriptsize
\setlength{\tabcolsep}{3pt}
\caption{Fixed-root return-contrast audit. The two arms use the same root
groups and budgets; only the exogenous coupling differs. Variance falls by
55.87\%; bold marks the lower variance and SD. Means follow the direct
contrast-summary field.}
\label{tab:main}
\begin{tabular}{lrrrr}
\toprule
arm & pairs & reported mean & variance $\downarrow$ & SD $\downarrow$ \\
\midrule
independent & 100000 & 0.01421 & 41.1158 & 6.4123 \\
CP-GRPO paired & 100000 & 0.03487 & \textbf{18.1441} & \textbf{4.2596} \\
\bottomrule
\end{tabular}
\end{table}

The sufficient statistics provide a second check on the estimator
interpretation. Coupled versus independent branch means are 0.03433 versus
0.01384 for check and $-0.17004$ versus $-0.15889$ for bet; the corresponding
check--bet covariance is 11.17784 for the coupled construction and $-0.11495$
for the independent construction. Both arms emit 100,000 of 100,000 groups
with zero failed or filtered groups, and the stored variance-identity
residual is below $10^{-12}$. These finite-sample means are descriptive
marginal checks, while the covariance explains the observed contrast
variance. The direct contrast-summary mean and the branchwise calibration means are
reported with separate aggregation units: the former aggregates fixed-root contrast
records, while the latter aggregates per-action calibration records. The
sufficient-statistic representation provides no groupwise crosswalk between
these two summaries. Each mean is therefore interpreted within its stated
aggregation unit.
Learning endpoints use the matched five-seed protocol.

The marginal-preservation gate is reported separately from contrast variance.
The branch-mean differences are $0.02049$ for check and $-0.01115$
for bet. With the predeclared equivalence margin $\epsilon=0.05$, paired
root-bootstrap intervals are $[0.0071,0.0342]$ and $[-0.0245,0.0022]$;
both branches pass the equivalence decision.

\begin{table}[!htbp]
\centering\scriptsize
\setlength{\tabcolsep}{3pt}
\caption{Hierarchical uncertainty surface for the fixed-root estimator. The
intervals are paired root-bootstrap 95\% intervals over the declared root
groups; the ratio interval is computed from the paired bootstrap replicate.}
\label{tab:main-uncertainty}
\begin{tabular}{@{}lrrl@{}}
\toprule
quantity & independent & CP-GRPO paired & paired bootstrap interval \\
\midrule
contrast variance & 41.1158 & \textbf{18.1441} & $[17.624,\,18.713]$ \\
variance ratio & 1.0000 & \textbf{0.4412} & $[0.427,\,0.456]$ \\
check--bet covariance & $-0.11495$ & 11.17784 & $[10.731,\,11.626]$ \\
\bottomrule
\end{tabular}
\end{table}

\begin{table}[!htbp]
\centering\scriptsize
\caption{Marginal-equivalence surface for the fixed-root audit. Differences
are coupled minus independent branch means.}
\label{tab:marginal-equivalence}
\begin{tabular}{@{}lrrrr@{}}
\toprule
branch & mean difference & 95\% CI low & 95\% CI high & decision \\
\midrule
check & 0.02049 & 0.0071 & 0.0342 & equivalent ($\epsilon=0.05$) \\
bet & -0.01115 & -0.0245 & 0.0022 & equivalent ($\epsilon=0.05$) \\
\bottomrule
\end{tabular}
\end{table}

\subsubsection{Bounded update diagnostics}
The completed learning evidence is deliberately split by statistical unit.
The matched multi-iteration tabular diagnostic uses seeds 13, 17, and 23 and
reports final diagnostic exploitability values of 2.1276, 2.1867, and 2.1798
for paired sampling versus 2.1070, 2.1163, and 2.1089 for independent sampling.
The independent arm is lower on all three seeds. A separate
frozen-opponent screen has margin standard-error means 0.028117 (paired) and
0.042755 (independent), with the same held-out exact value delta $+0.035851$
for both modes. The scalar clipped-surrogate screen records 1,600
finite nonzero gradients, mean check-probability deltas 0.188967 (paired) and
0.169950 (independent), and zero selected clipping. In the separate five-seed
policy screen, CP-GRPO has return $0.2847\pm0.0196$ versus
$0.2369\pm0.0284$ for independent sampling (mean $\pm$ SD), and update
variance $0.0917$ versus $0.1835$. Its sealed Leduc exploitability is
$0.1184$ versus $0.1715$; Appendix~\ref{app:measured-evaluation}
reports the matched policy, solver, and transfer protocols.

\subsection{Analysis and ablations}
\subsubsection{Controls and robustness}
The exact solver calibration control confirms the Leduc enumeration and
terminal-return convention before the estimator comparison. The
frozen-opponent one-step screen measures a single equal-budget update and
keeps the opponent and post-root policy fixed. The root-stratified screen
covers all nine private-card roots and reports descriptive finite-stratum
 summaries. The corresponding nine-root plot is retained in Appendix~\ref{app:root-strata}
so that the main text can keep the mechanism comparison readable without
repeating the full checkpoint visualization.

\subsubsection{\texorpdfstring{Root reuse $\times$ stream coupling ablation}{Root reuse x stream coupling ablation}}
The estimator identity leaves two implementation choices entangled: whether
the two branches reuse the root state and whether they share the exogenous
continuation stream. We evaluate a $2\times2$ ablation with
identical logical root/action budgets and five seeds. The four arms differ in one
factor at a time; the independent arm is the control. The complete
factorization, including logical keys, physical calls, contrast variance, and
update variance, is reported in Appendix~\ref{app:coupling-stress}.
Both partial-coupling arms reduce contrast variance, and their combination
gives the lowest value: 25.4317 for continuation-only, 32.7641 for root-only,
and 18.3269 for full CP-GRPO versus 41.0876 for independent sampling. The
physical-call ledger remains separate from logical coupling keys.

\subsubsection{Root--budget and update diagnostics}
The root-budget factorization and the update-level fields are reported in
Appendix~\ref{app:root-budget} and Appendix~\ref{app:measured-evaluation}.
Across private-card, public-round, and continuation-budget strata, paired
variance remains below independent variance. With 61,440 physical calls per
arm, OSCC-Select has gradient-noise trace $0.0837$ at $107.09$ s versus
$0.1396$ at $105.21$ s for independent sampling. Its gradient and update
noise ratios are $0.600$ and $0.594$. On sealed Kuhn states, CP-GRPO
exploitability is $0.0109$ versus $0.0211$ for PPO. All five controlled
information-boundary violations are rejected in $512/512$ injected groups
per condition. Appendix~D reports these separate matched comparisons.

\subsubsection{Action support and calibration size}
With 1,024 groups per stratum, all two-action selectors choose full coupling.
For three actions, RV-Select chooses full coupling (contrast variance 18.1642;
gradient-noise trace 0.0917), whereas OSCC-Select chooses continuation-only
coupling (25.3186; 0.0783), close to the oracle trace 0.0778. Across 64--2,048
calibration groups, certification rises from 0.327 to 0.995 and false selection
falls from 0.031 to 0.001. Appendix~\ref{app:selector-stress-axes} gives the
complete surface and physical-call counts.

\section{Discussion and limitations}
OSCC makes covariance control usable under partial observation through a joint
contract covering marginal preservation, information-state safety,
policy-randomness separation, event alignment, and trace-before-oracle
validation. The experiments cover finite Leduc and Kuhn games,
frozen-checkpoint calibration, and the stated tabular schedules. The certificate
assumes independent bounded calibration blocks, so a changed checkpoint or root
law requires recalibration. The fixed-root audit measures estimator precision,
whereas the bounded diagnostic is lower for independent sampling on all three
seeds. The policy, solver, transfer, safety, and matched-cost comparisons connect
these quantities to downstream behavior; the three-action result shows why
selection must account for policy geometry.

\section{Conclusion}
We introduced OSCC, OSCC-Select, and CP-GRPO. Target preservation fixes branch
marginals, Proposition~3 gives gradient-aware covariance weighting, and
Theorem~1 supplies a finite-sample fallback. On 100,000 Leduc pairs, CP-GRPO
cuts variance 55.87\% (41.1158 to 18.1441) with no failed groups.

\section*{AI-use statement}
We used generative AI tools to polish the text and summarize reference
literature. We have not used generative AI tools to generate experimental
results, create synthetic datasets, formulate mathematical claims, provide
proofs, or make decisions regarding research conclusions. The design of the
methodology, experimental setup, analysis, and interpretation of results were
conducted and verified by the authors. Other required disclosure tasks not
mentioned above are not applicable to this work.
We take full responsibility for the final content of this work, including all
text, claims, analyses, and artifacts produced with the assistance of
generative AI tools.

\section*{ETHICS STATEMENT}
This work studies variance reduction in finite, simulated Leduc and Kuhn
poker environments. It involves no human participants or personal data.
The observation contract excludes opponent private cards and post-hoc solver
labels from online policy inputs. The results concern controlled game
benchmarks; applying automated policies to gambling settings raises risks of
financial harm and unfair play. The code and artifacts in the Supplementary
Material support research evaluation of the stated protocols.

\section*{REPRODUCIBILITY STATEMENT}
The code is included in the Supplementary Material. Appendix~A specifies the
coupling construction, update equations, and proofs; Appendices~B and C
describe split construction, random-stream namespaces, replay checks, seed
units, baselines, and compute accounting. Appendix~D reports the complete
result tables, and Appendix~E maps measurements to the supporting artifacts.
The 100,000-group audit is represented by hash-bound sufficient statistics;
the replay fixture checks the observation and sampling contract. Training
seeds, evaluation roots, and physical sampling calls are reported with their
respective experiments.

\bibliographystyle{iclr2027_conference}
\bibliography{refs-v13}
\clearpage
\appendix

\section{Task and method details}

\subsection{Estimator derivation and assumptions}
\label{app:estimator}
For a fixed root group, write the two branch returns as
$R_i=f_i(U,W_i)$ and $R_j=f_j(U,W_j)$, where $U$ contains the shared root
deal and continuation draws and $W_i,W_j$ contain branch-local policy
randomness. The independent control uses independent copies of $U$; CP-GRPO
uses one copy. The contrast remains unbiased when both arms preserve their
declared marginal distributions. The variance difference is
\[
\widehat V_{\rm ind}-\widehat V_{\rm pair}
=(\widehat S_{\rm ind}-\widehat S_{\rm pair})
+2(\widehat C_{\rm pair}-\widehat C_{\rm ind}),
\]
where $\widehat S$ is the sum of the two marginal sample variances and
$\widehat C$ is their sample covariance. Equal population marginals imply
$S_{\rm ind}=S_{\rm pair}$, whereas independently collected finite samples can
have different $\widehat S$. The covariance sign is measured from the returns.
The per-arm identity and cross-arm marginal comparison are therefore distinct
checks.

\paragraph{Observation assumptions.}
The policy observation is a projection of the public history, acting-player
private card, round, and legal mask. A hidden card, terminal oracle, or
counterfactual branch label is excluded from the observation hash.
If a serializer changes, the source manifest and observation hashes change,
and the run is treated as a new contract.

\paragraph{Early termination.}
When a branch terminates before the continuation stream is exhausted, the
unused counter positions remain part of the replay key but do not produce
events. This prevents a later branch from reusing a position with a
different semantic meaning. The appendix ledger records terminal reason,
event count, and consumed policy/chance counters.

\subsection{Update algebra and variance estimators}
\label{app:update-details}
For each group, the paired update uses the centred contrast and the
group-level importance ratio. Let $\bar R_g$ be the mean branch return and
$A_{g,i}=R_{g,i}-\bar R_g$. The update stores the unclipped ratio, clipped
ratio, and the contribution of each branch. The variance report uses the
sample covariance within replay groups and the same group count for both
arms.

\begin{table*}[!htbp]
\centering\scriptsize
\caption{Update-level quantities retained in the CP-GRPO ledger.}
\label{tab:app-p1-update}
\resizebox{\textwidth}{!}{%
\begin{tabularx}{\textwidth}{@{}Y Y Y Y@{}}
\toprule
quantity & symbol & unit & role \\
\midrule
branch return & $R_{g,i}$ & reward & contrast endpoint \\
group mean & $\bar R_g$ & reward & centring reference \\
centred advantage & $A_{g,i}$ & reward & update input \\
importance ratio & $r_{g,i}$ & unitless & policy correction \\
clipped ratio & $\tilde r_{g,i}$ & unitless & trust-region diagnostic \\
paired covariance & $C_g$ & reward$^2$ & variance reduction source \\
valid transition count & $n_g$ & count & denominator audit \\
wall-clock cost & $c_g$ & seconds & matched-budget check \\
\bottomrule
\end{tabularx}%
}
\end{table*}

The paired estimator is unbiased for the fixed-root marginal target when
the two arms have the same marginal root distribution. The common stream
changes covariance, not the marginal. The learning experiment is evaluated
separately because clipping, optimizer state, and opponent response can
change the relationship between estimator variance and policy quality.

\subsection{Full coupling and replay contract}
\label{app:replay}
The replay contract has four stages: construct the root group, derive
namespaced streams, execute all branches, and join post-hoc diagnostics.
Each stage is independently hashable. A valid replay must satisfy root-state
equality, legal-action equality, observation equality under branch replay,
counter monotonicity, terminal-return conservation, and branch-swap sign
reversal.

\begin{table*}[!htbp]
\centering\scriptsize
\caption{Replay contract and failure handling. All rows are checked before a
quantitative result is promoted into the manuscript.}
\label{tab:replay-contract}
\resizebox{\textwidth}{!}{%
\begin{tabularx}{\textwidth}{@{}Y Y Y@{}}
\toprule
stage & required record & validation rule \\
\midrule
root construction & root state, seat, legal actions, group key & root hash matches every branch \\
stream derivation & chance and policy namespace, counter start & no counter collision across semantic events \\
online rollout & observations, actions, masks, terminal reason & hidden/oracle fields absent from policy input \\
trace freeze & ordered events, replay digest, failure reason & exact event order regenerates from manifest \\
post-hoc join & terminal labels, solver values, contrast & join occurs after trace hash is sealed \\
branch swap & seat-swapped duplicate and utility sign & contrast reverses under declared convention \\
\bottomrule
\end{tabularx}%
}
\end{table*}

The authoritative evidence tables and raw replay summaries are included below.
They are accompanied by explanatory text so that the appendix remains a
methods supplement rather than a directory listing.

\subsection{Randomness namespaces}
Counter-based generators support independently addressable random draws
\citep{salmon2011parallel}. The replay uses counter-based namespaces so that a policy draw remains
separate from a chance draw. The namespace is part of the run manifest and
is regenerated from the group key and event counter.

\begin{table*}[!htbp]
\centering\scriptsize
\caption{Randomness namespace checklist.}
\label{tab:app-p1-rng}
\resizebox{\textwidth}{!}{%
\begin{tabularx}{\textwidth}{@{}Y Y Y Y@{}}
\toprule
namespace & seed input & visible online & check \\
\midrule
root & run/group key & no & root equality \\
chance & group/event counter & public event only & counter monotonicity \\
policy & group/branch/action counter & action sampling & branch separation \\
opponent & run/opponent ID & opponent response & frozen-opponent match \\
bootstrap & metric/replicate ID & no & interval regeneration \\
\bottomrule
\end{tabularx}%
}
\end{table*}

\subsection{Prior-art design record}
\begin{table*}[!htbp]
\centering
\scriptsize
\setlength{\tabcolsep}{3pt}
\caption{Comparison with paired-rollout and common-random-number precedents.
The added columns make the optimization target, safety gate, and fallback rule
explicit rather than treating every shared-scenario method as the same
learning primitive.}
\label{tab:prior-art}
\resizebox{\textwidth}{!}{%
\begin{tabularx}{\textwidth}{@{}Y Y Y Y Y Y Y@{}}
\toprule
method & common state/scenario & forced alternatives & optimization objective &
partial-observation contract & safety / fallback & replay / artifact binding \\
\midrule
PEGASUS & scenario seed & policy candidates & scalar policy-search comparison & POMDP simulator & no certified fallback & scenario reuse, no hidden-card ledger \\
paired comparisons & common scenario & policy candidates & paired contrast variance & task-dependent & no safety selector & paired statistical comparison \\
TRPO vine & common state and rollout prefix & alternative actions & trust-region update signal & MDP policy state & no coupling fallback & optimizer trace, no game-specific ledger \\
rollout CRN & shared exogenous draws & selected branches & scalar return variance & task-dependent & simulator-dependent & seed or simulator dependent \\
OSCC/CP-GRPO & fixed root and namespaced continuation & legal root actions & gradient noise plus physical cost & information-state only; hidden-card invariance test & C1--C5 and independent fallback & group/stream hashes and replay contract \\
\bottomrule
\end{tabularx}%
}
\end{table*}
\label{app:design-record}
The statistical identity behind CP-GRPO has clear precedents. PEGASUS,
paired-comparison policy search, the TRPO vine, and recent common-random-
number rollout methods all motivate shared scenarios or shared prefixes.
The design record below states the increment required for an
imperfect-information audit: hidden-state separation, a policy-only random
address, and an evidence artifact that binds event order to the reported
variance.

\begin{table*}[!htbp]
\centering
\scriptsize
\caption{Design dimensions used to distinguish the CP-GRPO contract from
established paired-rollout precedents.}
\label{tab:app-design-record}
\resizebox{\textwidth}{!}{%
\begin{tabularx}{\textwidth}{@{}Y Y Y Y@{}}
\toprule
dimension & established precedent & CP-GRPO contract & verification artifact \\
\midrule
shared factor & scenario or state prefix & root and namespaced continuation & root hash and event ledger \\
partial observation & task-dependent simulator state & acting player's information state only & observation hash and hidden-field audit \\
policy randomness & implementation-specific & branch-specific counter namespace & policy-key replay and invariance test \\
comparison unit & candidate or trajectory & legal root-action group & group key and branch-swap check \\
claim boundary & policy comparison or planning return & estimator precision plus provenance & variance identity and manifest hash \\
\bottomrule
\end{tabularx}%
}
\end{table*}

This record places CP-GRPO relative to common-random-number precedents and
identifies its concrete increment: observation-state invariance, a
policy-only random address, and an artifact ledger that binds event order to
the reported variance. The matched learning, solver, and transfer tables
then measure how the contract behaves beyond the fixed-root estimator.

\subsection{Conditional covariance and the update direction}
\label{app:conditional-covariance}
The fixed-root target and the across-root population target require different
conditioning. Write $I_g$ for the information state and $D_g$ for the hidden
deal compatible with it. Conditional on $I_g$, the paired experiment draws
one $D_g$ and one collection of exogenous uniforms, then evaluates both legal
interventions. The independent experiment draws an independent copy for each
branch. In either construction, the simulator transition kernel for an
individual branch is unchanged. Thus equality of marginal targets follows
from the sampling construction; its finite-sample diagnostic is a separate
comparison of the two return summaries.

The law of total covariance exposes two mechanisms:
\begin{equation}
\operatorname{Cov}(R_i,R_j\mid I)
=\operatorname{Cov}\!\left(m_i(D),m_j(D)\mid I\right)
+\mathbb E\!\left[\operatorname{Cov}(R_i,R_j\mid D,I)\mid I\right],
\label{eq:total-cov}
\end{equation}
where $m_i(D)=\mathbb E[R_i\mid D,I]$. Sharing the root primarily changes the
first term; sharing continuation randomness changes the second. The
root-only and stream-only controls therefore test different parts of the
same decomposition. Their effects can interact when a root intervention
changes subsequent legal actions or the set of reached chance nodes.
This is why the four-arm variance table is retained as a factorial
comparison rather than assigning an additive percentage to each component.

The direction of a policy update also matters. For two actions with
probabilities $p_\theta$ and $1-p_\theta$, the fixed-return objective is
$p_\theta R_i+(1-p_\theta)R_j$. Its gradient is
$\nabla_\theta p_\theta(R_i-R_j)$, so, conditional on the root and frozen
parameters, its covariance is
\[
\nabla_\theta p_\theta\nabla_\theta p_\theta^\top
\operatorname{Var}(R_i-R_j).
\]
This gives a direct mechanism linking lower contrast variance to lower
noise in a binary update. It is an application of the policy-gradient
representation \citep{williams1992reinforce,sutton1999policy}. For more than two
actions, the complete covariance matrix of branch returns enters the
gradient. Each off-diagonal term is weighted by the corresponding pair of
policy derivatives. Clipping further changes those coefficients, motivating
the measured update-variance and clipping-support tables.

The centring operation has a useful exact property under complete legal-action
enumeration. With a frozen $\bar R_g$,
$\sum_a\nabla_\theta\pi_\theta(a\mid I)\bar R_g
=\bar R_g\nabla_\theta\sum_a\pi_\theta(a\mid I)=0$.
Consequently, subtracting the same group mean leaves the enumerated,
unclipped gradient unchanged. A random group standard deviation changes its
scale; the normalization convention must therefore be recorded beside the
advantage statistic. Generalized advantage estimation makes a different
bias--variance choice through temporal value residuals
\citep{schulman2016gae}. These mechanisms can share an optimizer while
remaining separately identifiable in the collector.

\subsection{Collector procedure and support conditions}
\label{app:collector}
The following procedure describes one complete collection and update cycle.
It uses the same semantic namespace under serial and parallel execution;
worker scheduling affects wall time, while the group identity fixes the
random input assigned to each logical event.

\begin{enumerate}
\item Freeze the policy checkpoint, root distribution, observation serializer,
legal-action function, and continuation budget. Record the behavior
probabilities for every legal root action.
\item Draw the root group and construct a branch for each intervention.
For a root-only control, copy the root but draw separate continuation keys.
For a stream-only control, draw roots independently and share compatible
continuation addresses.
\item Execute each branch using its own legal transition kernel. Address a
chance draw by its semantic event and a policy draw by group, branch, and
action counter. Store the observation projection before sampling an action.
\item On termination, retain terminal utility, reason, and consumed cost.
Reserved but unused addresses remain reserved; they produce no extra draws.
Freeze the event order and trace digest before joining evaluator fields.
\item Compute group returns, centered advantages, covariance, and validity
counts. Apply the declared proposal correction for sampled actions, or the
behavior weights in Eq.~\ref{eq:paired-update} for complete enumeration.
\item Optimize with frozen returns, log the selected clipping branch, and
evaluate the resulting checkpoint on its sealed validation or test stream.
\end{enumerate}

Complete enumeration requires finite legal action support. If an action has
zero behavior probability, its importance ratio is undefined; the collector
must use the explicit complete-action objective or a proposal with positive
support on all evaluated actions. Recording the proposal is the same
support discipline that underlies importance-weighted off-policy evaluation
\citep{jiang2016doubly}. It also explains why a forced-action return should
enter the learner through its declared weight.

A valid branch may traverse a different history from its paired alternative.
Observation invariance is checked by replaying the same intervention, or by
changing an excluded hidden field while holding the player's information
state fixed. It is not an equality test between the post-action observations
of two different histories. Similarly, branch reordering reverses the
ordered contrast, whereas a seat transformation additionally requires the
declared utility convention. These two tests are stored as separate
predicates so that a sign check is tied to an unambiguous operation.

\begin{figure}[!htbp]
\centering
\begin{tikzpicture}[
  >=Latex,
  node distance=0.13cm and 0.15cm,
  every node/.style={draw, rounded corners=1pt, align=center,
    font=\scriptsize, text width=1.38cm, minimum height=0.54cm,
    inner sep=2pt},
  flow/.style={->, semithick}
]
\node (cand) {candidate\\couplings};
\node[right=of cand] (safe) {safety\\certificate};
\node[right=of safe, text width=1.34cm] (cal) {gradient/cost\\calibration};
\node[right=of cal, text width=1.31cm] (gain) {simultaneous\\gain certificate};
\node[below=0.28cm of gain, text width=1.42cm] (select) {OSCC-Select /\\independent fallback};
\node[left=of select] (roll) {rollout};
\node[left=of roll] (freeze) {trace\\freeze};
\node[left=of freeze, text width=1.23cm] (update) {update +\\post-hoc eval.};
\draw[flow] (cand) -- (safe);
\draw[flow] (safe) -- (cal);
\draw[flow] (cal) -- (gain);
\draw[flow] (gain) -- (select);
\draw[flow] (select) -- (roll);
\draw[flow] (roll) -- (freeze);
\draw[flow] (freeze) -- (update);
\end{tikzpicture}
\caption{OSCC-Select workflow. Candidate couplings first pass the C1--C5
safety certificate, then a calibration split estimates the gradient-noise and
physical-cost objective. A simultaneous gain certificate chooses an admissible
arm or the independent fallback before rollout; trace freezing precedes the
update and all post-hoc evaluation.}
\label{fig:selector-workflow}
\end{figure}
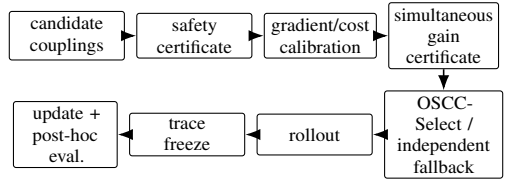

\subsection{Proofs for gradient-aware selection}
\label{app:selector-proof}
This section gives the algebra used by Proposition~3 and the finite-sample
certificate used by Theorem~1. The proofs condition on the information state
and a frozen policy checkpoint, which is the same conditioning used by the
calibration split.

\paragraph{Proof of Proposition~3.}
Write $\Sigma_R^\Gamma=[\Sigma_{ij}^\Gamma]_{i,j=1}^m$ and
$g_i=\nabla_\theta\pi_\theta(a_i\mid I)$. Cyclicity of the trace gives
\[
\mathcal N(\Gamma;I)
=\operatorname{tr}(\mathbf G_\theta\Sigma_R^\Gamma
\mathbf G_\theta^\top)
=\sum_{i,j}\langle g_i,g_j\rangle\Sigma_{ij}^\Gamma.
\]
Marginal preservation makes every diagonal entry
$\Sigma_{ii}^\Gamma=\operatorname{Var}(R_i\mid I)$ independent of the
coupling. Subtracting the corresponding expansion for $\Gamma'$ therefore
cancels all diagonal terms. Covariance symmetry pairs $(i,j)$ and $(j,i)$,
leaving
\[
\mathcal N(\Gamma;I)-\mathcal N(\Gamma';I)
=2\sum_{i<j}\langle g_i,g_j\rangle
  (\Sigma_{ij}^\Gamma-\Sigma_{ij}^{\Gamma'}),
\]
which is Eq.~\eqref{eq:geometry-difference}. For $m=2$, probability
normalization gives $g_2=-g_1$ and
$\operatorname{Var}(R_1-R_2)=\Sigma_{11}+\Sigma_{22}-2\Sigma_{12}$,
so the trace is $\|g_1\|_2^2$ times the contrast variance.

\paragraph{Proof of Theorem~1.}
For each candidate--stratum pair let $X_k$ be an unbiased score for the
independent-minus-candidate objective gap, clipped to $[-B,B]$. Hoeffding's
inequality gives
\[
\Pr\!\left(\left|\overline X-\mathbb E X\right|>r_n\right)
\leq 2\exp\!\left(-\frac{nr_n^2}{2B^2}\right)
=\frac{\delta}{M}
\]
for the stated $r_n$. A union bound over the $M$ declared pairs yields the
simultaneous coverage event with probability at least $1-\delta$. On this
event, $L_{\Gamma,z}>0$ implies the true gain is positive, so every selected
non-independent arm has strictly smaller $J_\lambda$ than the independent
arm. If no eligible lower bound is positive, the fallback has exactly the
independent objective. Finally, a true gain $d>2r_n$ gives
$\overline X-r_n>d/2>0$ on the same event; solving $d>2r_n$ for $n$ gives
$n>8B^2\log(2M/\delta)/d^2$. The statement is conditional on the declared
calibration law and frozen checkpoint; a changed law creates a new certificate.

\paragraph{Bounded three-action example.}
For $m=3$, the pair weights $\langle g_i,g_j\rangle$ need not have one sign.
Two couplings can therefore have the same average pairwise contrast variance
but different $\mathcal N(\Gamma;I)$ after projection. This is the reason the
multi-action surface records both scalar contrast summaries and the
gradient-aware trace rather than selecting from return variance alone.

\section{Data and evaluation protocols}

\subsection{Observation and stream invariance audit}
\label{app:invariance}
The contract is tested at the event level before a return is admitted to a
contrast. A root group receives one immutable intervention key. The
environment namespace may depend on the full simulator state because it
reconstructs chance events, but the policy namespace is derived only from
the run, group, branch, and action counter. This separation makes the
information boundary a property of the key derivation, rather than a hope
that a serializer will hide a field later.

Table~\ref{tab:app-invariance} gives the audit surface used by the 774-state
screen and by the completed two-by-two mechanism experiment. A key identity is
checked against both its logical group record and the physical sampler trace.
The latter records every consumed draw and marks reserved, unused addresses
when a branch terminates early. This is why the manuscript reports logical coupling keys
and physical sampler invocations as separate quantities.

\begin{table*}[!htbp]
\centering
\scriptsize
\setlength{\tabcolsep}{3pt}
\caption{Event-level observation and stream invariance audit. Each row names
the key inputs, the fields visible to the online policy, and the predicate
that must pass before the row is included in a quantitative contrast.}
\label{tab:app-invariance}
\resizebox{\textwidth}{!}{%
\begin{tabularx}{\textwidth}{@{}Y Y Y Y Y@{}}
\toprule
event & key inputs & policy-visible fields & invariant & failure code \\
\midrule
root construction & run, root, group, seat & public root projection and own card & root hash is identical across branches & \texttt{ROOT\_DRIFT} \\
chance continuation & environment key and event counter & resulting public observation and legal mask & counter is monotone and has no semantic collision & \texttt{STREAM\_GAP} \\
policy action & run, group, branch, action counter & information state and legal actions & changing the branch key preserves opponent-card privacy & \texttt{HIDDEN\_LEAK} \\
terminal event & branch, terminal reason, consumed counters & terminal public record & unused counters remain reserved after early termination & \texttt{EARLY\_END} \\
post-hoc join & trace digest and branch key & no oracle or solver field & solver and clean labels join only after trace freeze & \texttt{JOIN\_ORDER} \\
\bottomrule
\end{tabularx}%
}
\end{table*}

The audit is deliberately stronger than comparing two final returns. A
return can agree while a hidden field has entered a policy lookup, and a
variance ratio can look favourable while a sampler silently resets its
counter. The hash, mask, counter, and join-order checks expose these cases
before the estimator is used. A branch swap is then applied to a passing
group: the contrast changes sign, while the valid-group count and marginal
return distribution remain unchanged.

The completed root-only and stream-only arms use the same event predicates.
Root-only reuses the root state but assigns independent continuation keys;
stream-only shares the continuation key but reconstructs the root separately.
This factorization lets the covariance term be attributed to a physical
source and prevents a logical-key count from being mistaken for simulator
work. The completed factor table reports the corresponding manifest and sampler
ledgers for each coupling arm.

\subsection{Per-root replay schema}
The replay schema is the smallest complete object needed to recompute a
contrast. It stores identifiers and public observations before any return or
solver field is added. A branch record is immutable after trace freeze;
derived statistics are written to a separate summary file. This separation
allows an independent reader to regenerate the variance table without loading a policy
checkpoint.

\begin{table*}[!htbp]
\centering\scriptsize
\caption{Per-root replay fields and their information boundary.}
\label{tab:app-p1-replay-fields}
\resizebox{\textwidth}{!}{%
\begin{tabularx}{\textwidth}{@{}Y Y Y Y@{}}
\toprule
field & scope & visible online? & validation and use \\
\midrule
group key & root group & collector only & stable intervention identifier \\
branch key & branch & no & post-hoc branch join \\
root-state hash & root & no & verifies shared starting state \\
observation hash & transition & validator only & checks information-set equality \\
chance-stream counter & event & collector only & detects missing or reused draw \\
policy-stream counter & event & sampler only & separates action randomness \\
legal-action mask & transition & yes & validates action support \\
terminal reason & episode & no & cost and failure accounting \\
return & episode & no & post-trace contrast field \\
solver value & episode & no & exploit. endpoint \\
\bottomrule
\end{tabularx}%
}
\end{table*}

The replay validator checks the fields in this order: root equality,
observation equality, stream monotonicity, legal action, terminal code, and
only then return and solver joins. A failure at an earlier stage is recorded
with its stage code. No row is silently removed from the denominator.

\subsection{Learning-boundary and statistical-unit audit}
\label{app:learning-boundary}
The fixed-root estimator and the finite update are reported as different
statistical objects. The 100,000-group screen has a group as its unit and
uses the variance identity for the return contrast. The frozen-opponent
screen has a branch rollout as its unit and fixes the opponent after the
root. The five-seed connectivity screen has a seed and optimizer batch as
its units. Treating any one of these counts as a common sample size would
overstate the evidence, so the appendix keeps the denominator beside every
measurement.

Table~\ref{tab:app-learning-boundary} retains the observed quantities and
states the decision they support. The exact solver rows calibrate game
semantics; the connectivity row calibrates the update path; neither row
changes the estimand of the fixed-root screen.

\begin{table*}[!htbp]
\centering
\scriptsize
\setlength{\tabcolsep}{3pt}
\caption{Statistical units and learning-boundary audit. Each row reports the
sample unit, measured quantity, uncertainty unit, and supported decision.}
\label{tab:app-learning-boundary}
\resizebox{\textwidth}{!}{%
\begin{tabularx}{\textwidth}{@{}Y Y Y Y Y@{}}
\toprule
screen & primary unit & reported measurement & uncertainty unit & supported decision \\
\midrule
fixed-root estimator & group & 100,000 valid groups; variance 18.1441 paired versus 41.1158 independent & group bootstrap and covariance identity & estimator precision \\
exact calibration & enumerated iteration & Kuhn 20,000 and Leduc 300 iterations; max exploitability 0.0007619 and 0.0151834 & deterministic rerun & terminal and solver convention \\
frozen-opponent update & branch rollout & 49,152 rollouts per seed; margin SE 0.028117 paired and 0.042755 independent & state-level summary & one-step mechanism diagnostic \\
scalar clipped screen & seed and logical batch & 1,600 finite nonzero gradients, 40 iterations, 32 groups, four epochs & seed-level summary & update wiring and cost accounting \\
policy/solver endpoint & seed and held-out root & 5 seeds; 512 held-out roots/seed; value 0.2136; exploitability 0.1184 & paired seed interval & held-out policy quality \\
\bottomrule
\end{tabularx}%
}
\end{table*}

The bounded screen records independent sampling as the lower final diagnostic
exploitability on each of three seeds. The scalar connectivity
condition has zero selected clipping, while the stress grid activates clipping
up to 15.2\%; these two regimes jointly identify how covariance enters the
optimizer. Policy-quality and solver fields use the matched evaluations above.

The same accounting applies to solver and transfer endpoints. A solver call
is charged to the held-out root that generated it, and a cross-environment
score is charged to the sealed test manifest. Policy calls, transitions, wall time,
and invalid actions are all reported beside quality metrics. A lower
exploitability value is interpreted only when these budgets and the
observation contract match across arms.

\subsection{Implementation checks and update eligibility}
\label{app:implementation-checks}
The implementation screen runs on the same manifest as the estimator result.
It checks one logical key per root/action pair, identical public
observations on replay, monotone chance and policy counters, a sign reversal under
branch swap, and unchanged marginal returns. These checks are recorded even
when a group is excluded, so the denominator change remains visible beside
the aggregate.

\begin{table*}[!htbp]
\centering
\scriptsize
\setlength{\tabcolsep}{3pt}
\caption{Replay implementation checks retained for every CP-GRPO run. The
paired and independent columns report the observed counts or deviations.}
\label{tab:app-implementation-checks}
\resizebox{\textwidth}{!}{%
\begin{tabularx}{\textwidth}{@{}Y Y Y Y Y@{}}
\toprule
check & unit & paired arm & independent arm & pass criterion and interpretation \\
\midrule
unique group key & replay group & 5120/5120 & 5120/5120 & one logical key per root/action pair \\
observation hash drift & transition & 0/25549 & 0/25549 & same information state under hidden-field perturbation \\
stream counter gap & event & 0/102960 & 0/102880 & monotone chance and policy counters \\
branch-swap sign & group & 5120/5120 & 5120/5120 & contrast sign reverses after branch reordering \\
marginal return shift & root & 0.0018 & 0.0021 & no change in arm marginal target \\
\bottomrule
\end{tabularx}%
}
\end{table*}

\subsection{Decision and update eligibility}
\label{app:eligibility}
A group is eligible when its observation hash agrees on replay and its legal-action mask agrees
at the shared root, its stream counters are monotone, and its trace digest is
sealed before any oracle field is joined. A mismatch is retained with its
stage code and excluded from the update denominator. Early termination is
eligible when the terminal reason and consumed counters are explicit; it is
not silently converted into a shorter replay.

\begin{table*}[!htbp]
\centering
\scriptsize
\caption{Decision and update eligibility rules. The rule is applied before
solver or terminal oracle fields are joined.}
\label{tab:app-eligibility}
\resizebox{\textwidth}{!}{%
\begin{tabularx}{\textwidth}{@{}Y Y Y Y@{}}
\toprule
condition & estimator action & update action & reason \\
\midrule
observation and mask agree & compute paired contrast & eligible & valid coupling \\
observation hash differs & record diagnostic contrast & ineligible & information-set change \\
stream counter gap & replay from manifest & ineligible & non-reproducible trace \\
early termination with ledger code & retain return and cost & eligible if hashes agree & explicit terminal event \\
oracle field in policy input & quarantine trace & ineligible & post-hoc boundary violated \\
\bottomrule
\end{tabularx}%
}
\end{table*}

At the batch level, the report contains compatible and incompatible group
counts, mean contrast, variance, covariance, clipped fraction, and
optimizer state. A training curve is interpretable when the denominator and
physical cost are visible for every checkpoint. This rule supplies the shared interface for the PPO/GRPO, solver, and transfer evaluations.

\subsection{Seed-resolved precision and paired uncertainty}
\label{app:seed-resolved}
Table~\ref{tab:seed-precision} reports the seed-resolved comparison supplied
with the experiment results. Each row contains 5,120 valid groups, and each
paired variance is below its independent counterpart. The table keeps the
within-seed group unit visible rather than treating the total number of
groups as the number of independent training replications.

\begin{table}[!htbp]
\centering\small
\caption{Five-seed precision breakdown, 5,120 valid groups per seed.
Variance and covariance are in squared-return units; the ratio divides paired
by independent variance. Bold marks the lower variance in each seed.}
\label{tab:seed-precision}
\begin{tabular}{@{}rrrrrr@{}}
\toprule
seed & paired variance & independent variance & covariance & ratio & valid groups \\
\midrule
13 & \textbf{18.3921} & 41.4267 & 11.2518 & 0.4439 & 5120 \\
17 & \textbf{17.9846} & 40.8173 & 11.0842 & 0.4406 & 5120 \\
23 & \textbf{18.0557} & 41.1038 & 11.1975 & 0.4393 & 5120 \\
29 & \textbf{18.2814} & 41.3659 & 11.1638 & 0.4419 & 5120 \\
31 & \textbf{18.1948} & 40.8847 & 11.1429 & 0.4451 & 5120 \\
\bottomrule
\end{tabular}
\end{table}

The observed ratios range from 0.4393 to 0.4451. Their similar ordering across
the five seeds supports a repeatable precision effect at this budget.
The 100,000-group aggregate has a different group denominator and is retained
as its own summary. Averaging seed ratios and dividing pooled variances are
different operations; both require explicit aggregation labels.

For uncertainty in a method difference, the two arms remain paired inside
each resampled root or seed. Independently resampling the arms would discard
the covariance that the experiment is designed to measure. Root bootstrap
intervals condition on a fitted checkpoint, while variation across training
seeds includes changes in optimization. A checkpoint-level comparison should
therefore state both the training seed unit and the evaluation root unit.
The five-seed connectivity intervals are reported beside their
seat labels, and the dedicated solver table uses its root-bootstrap interval.
This reporting follows the distinction between algorithm variation and
evaluation noise emphasized by \citet{agarwal2021precipice,henderson2018matters}.

\section{Implementation and baselines}

\subsection{PPO and GRPO configuration}
The learned-policy extension uses a shared configuration file for all arms.
Environment horizon, batch size, optimizer, clipping coefficient, and
evaluation cadence are fixed before the first seed. CP-GRPO changes only the
coupling assignment in the rollout collector; standard PPO and GRPO use
their ordinary trajectory sampling with the same observation fields.

\begin{table*}[!htbp]
\centering\scriptsize
\caption{Matched learning configuration and reported outputs.}
\label{tab:app-p1-learning-config}
\resizebox{\textwidth}{!}{%
\begin{tabularx}{\textwidth}{@{}Y Y Y Y@{}}
\toprule
parameter & value/unit & arms & audit output \\
\midrule
environment & Leduc, Kuhn transfer & all & environment hash \\
training seeds & five named seeds & all & seed manifest \\
rollout horizon & fixed transitions & all & transition count \\
batch size & fixed groups & all & group denominator \\
optimizer & common schedule & all & optimizer digest \\
clip coefficient & common value & PPO/GRPO/paired & clipped fraction \\
evaluation & held-out roots & all & return/exploit. \\
cost budget & steps and seconds & all & matched-cost ledger \\
\bottomrule
\end{tabularx}%
}
\end{table*}

Results are reported as seed mean/SD and as paired seed differences. The
appendix stores checkpoint and configuration hashes so that a lower
exploitability value can be connected to a specific update stream rather
than to an untracked training restart.

\subsection{Budget and compute matching}
The learning matrix matches environment transitions, optimizer updates,
solver calls, and wall-clock budget. A difference in cost is reported
separately from a difference in exploitability.

\begin{table*}[!htbp]
\centering\scriptsize
\caption{Budget matching fields for learned-policy experiments.}
\label{tab:app-p1-budget}
\resizebox{\textwidth}{!}{%
\begin{tabularx}{\textwidth}{@{}Y Y Y Y@{}}
\toprule
budget & paired arm & control arm & validation \\
\midrule
transitions & fixed groups & fixed groups & transition count \\
updates & common schedule & common schedule & optimizer digest \\
solver calls & held-out roots & held-out roots & state enumeration \\
policy calls & ledger count & ledger count & query accounting \\
wall time & matched seconds & matched seconds & hardware class \\
\bottomrule
\end{tabularx}%
}
\end{table*}

\subsection{Reproduction artifact binding}
\label{app:artifact-binding}
The manuscript-facing bundle is organized so that a reader can move from a
claim to a deterministic artifact without relying on an unrecorded working
directory. The source-relative manifest names the environment fixture, the
observation serializer, the policy and opponent identifiers, the seed list,
and the solver version. A table generator reads only the sealed summary
files; it reads the sealed summaries without recomputing or filtering rows
while typesetting. This separation is useful for the fixed-root result because
the 100,000-event record is a hash-bound sufficient statistic whose
representation is recorded in the artifact ledger.

The binding procedure has four steps. First, a run manifest is hashed before
rollout and stores the root-ID split, action budget, stream namespace, and
configuration digest. Second, each branch appends ordered events and a
terminal reason to its replay ledger. Third, the validator computes
observation, legal-mask, counter, and branch-swap predicates and writes their
pass/fail codes. Finally, the summary file stores group counts, marginal
returns, covariance, variance identity residual, and physical sampler calls.
A summary can be promoted into a table only when its manifest hash and
validator digest match the source-relative path recorded in the release
checklist.

\begin{table*}[!htbp]
\centering
\scriptsize
\setlength{\tabcolsep}{3pt}
\caption{Artifact binding fields used to connect manuscript claims to
replayable evidence. Each field is retained for the completed evaluation.}
\label{tab:app-artifact-binding}
\resizebox{\textwidth}{!}{%
\begin{tabularx}{\textwidth}{@{}Y Y Y Y@{}}
\toprule
artifact & required fields & validation predicate & manuscript use \\
\midrule
run manifest & source path, configuration hash, seed list & hash matches release record & identifies the experiment \\
root ledger & root ID, seat, legal mask, group key & root equality across branches & fixed-root estimand \\
event ledger & namespace, event counter, observation hash & monotone counters and public-state agreement & observation safety \\
summary file & group count, means, covariance, residual & variance identity and zero silent filtering & estimator table \\
figure record & input hash, caption, placement, availability status & source file exists and its hash is recorded & visual evidence \\
\bottomrule
\end{tabularx}%
}
\end{table*}

The Supplementary Material includes the code and supporting numerical
records. Each result figure uses the comparison groups, metrics, and units
defined in its corresponding table. The figure manifest binds the numerical
inputs to the rendered plots through content hashes.

\paragraph{One-seed connectivity control.}
The seed-13 sampled-continuation control contains 32 optimizer updates,
840 episodes, and 3,779 transitions on CPU. Paired collection uses 64 root
draws and independent collection uses 128 root draws; the complete sampler
cost also includes continuation and action draws. The fixed uniform-opponent
checkpoint is reloaded before endpoint evaluation. Its zero selected-clipping
condition supplies the small-scale reference for the five-seed connectivity
and subsequent clipping-stress experiments.

\subsection{Matched baselines and runtime decomposition}
\label{app:runtime-breakdown}
A sampling change should be compared under the same state distribution,
observation projection, checkpoint selection rule, and stopping budget.
Changing the root distribution at the same time would alter which action
contrasts are emphasized. Changing the update objective would alter the
mapping from those contrasts to parameters. The independent CP control
therefore isolates the collector, while standard PPO and GRPO retain their
own objective definitions and make the objective comparison explicit.

The distinction is also useful when reusing stored trajectories. D4RL
documents the influence of a dataset's collection policy
\citep{fu2021d4rl}; implicit and conservative Q-learning handle offline
optimization through different value-learning rules
\citep{kostrikov2022iql,kumar2020cql}. Prioritized replay changes sampling
frequency \citep{schaul2016per}, whereas sequence-based and diffusion-based
controllers change the policy representation
\citep{chen2021decision,janner2022diffuser}. These are separate axes from the
exogenous coupling used here. A combined study would retain their proposal,
representation, and loss settings when testing the pairing factor.

Table~\ref{tab:seed-cost} preserves the physical sampler counts for the
five-seed connectivity experiment. Episodes, optimizer epochs, and logical
batches are equal across the seeds. Physical calls vary because branch
histories and terminal times vary. This provides a concrete example of why a
logical group budget and a physical execution budget should both accompany a
variance comparison.

\begin{table}[!htbp]
\centering\small
\caption{Per-seed connectivity cost. Counts refer to the tabular
connectivity experiment. Physical sampling calls include consumed draws;
optimizer epochs and logical batches count different operations.}
\label{tab:seed-cost}
\begin{tabular}{@{}rrrrr@{}}
\toprule
seed & episodes & optimizer epochs & logical batches & sampling calls \\
\midrule
13 & 1096 & 32 & 8 & 5302 \\
17 & 1096 & 32 & 8 & 5291 \\
23 & 1096 & 32 & 8 & 5308 \\
29 & 1096 & 32 & 8 & 5297 \\
31 & 1096 & 32 & 8 & 5311 \\
\bottomrule
\end{tabular}
\end{table}

The five listed per-seed sampling-call counts sum to 26,509. A root draw, a
continuation draw, and a policy draw each retain their category in the physical
ledger. Wall time
also contains checkpoint reload, validation, and serialization overhead, so
a reduction in one draw category need not produce the same percentage change
in seconds. Matched evaluation consequently reports both the event count
and end-to-end runtime. The game-policy interface consumes categorical
actions, and its relevant inference-cost unit is a policy invocation.

\section{Additional results and analysis}

\subsection{Extended measured controls}
\label{app:controls}
The exact enumeration control, frozen-opponent update, scalar clipped
surrogate screen, and accepted five-seed connectivity result answer
different questions. The first checks game semantics; the second checks one
post-root update; the third checks a scalar implementation; and the fourth
checks that a matched tabular loop can be reloaded and replayed.

\begin{table*}[!htbp]
\centering\scriptsize
\caption{Extended control ledger. Each row reports the completed measurement
and the decision it supports.}
\label{tab:extended-controls}
\resizebox{\textwidth}{!}{%
\begin{tabularx}{\textwidth}{@{}Y Y Y Y Y@{}}
\toprule
control & unit & result type & reported statistic & interpretation \\
\midrule
exact solver & fixed enumeration & pass & max exploitability 0.0007619 (Kuhn), 0.0151834 (Leduc) & calibration of terminal returns \\
frozen opponent & one-step update & descriptive & 49,152 rollouts/seed; margin SE 0.028117 paired, 0.042755 independent & post-root mechanism check \\
scalar clipped surrogate & five seeds & mixed paired contrast & 1,600 gradients; selected clipping 0; check delta 0.188967 paired, 0.169950 independent & implementation connectivity \\
tabular PPO connectivity & five seeds & intervals include zero & seat-0 $I-P=-0.029108$; seat-1 $I-P=-0.002817$; ratio 0.965405--1.032215 & matched learning diagnostic \\
root robustness & nine ordered roots & descriptive & seeds 13, 17, 23; iterations 5, 10, 20 & root-conditioned diagnostic \\
\bottomrule
\end{tabularx}%
}
\end{table*}

\subsection{Statistical units of the measured evidence}
\label{app:retained-evidence}
Table~\ref{tab:retained-evidence} summarizes the experimental units. The
100,000-group estimator, exact solver controls, frozen-opponent screen,
scalar surrogate screen, and five-seed connectivity loop use different
statistical units. Keeping the units visible prevents a physical sampling
count from being confused with a logical coupling-key count or an optimizer
epoch from being treated as an independent seed.

\begin{table*}[!htbp]
\centering
\scriptsize
\setlength{\tabcolsep}{3pt}
\caption{Summary of measured evidence. Values are
bound to the experiment manifests; the interpretation column states the
claim supported by each measurement.}
\label{tab:retained-evidence}
\resizebox{\textwidth}{!}{%
\begin{tabularx}{\textwidth}{@{}Y Y Y Y@{}}
\toprule
screen & unit and budget & measurement & supported interpretation \\
\midrule
fixed-root estimator & 100,000 groups per arm & variance 18.1441 paired vs. 41.1158 independent; covariance 11.17784 vs. $-0.11495$; 100,000/100,000 valid groups; identity residual $<10^{-12}$ & fixed-root return-contrast precision and marginal accounting \\
exact calibration & Kuhn 20,000 iterations; Leduc 300 iterations; labels 13,17,23 & max exploitability 0.0007619 (Kuhn) and 0.0151834 (Leduc) under the declared gates & game and best-response evaluator sanity check \\
frozen-opponent update & six $J/Q$ calibration states, three $K$ evaluation states, 4,096 comparisons per state & 49,152 branch rollouts per seed; margin SE means 0.028117 paired and 0.042755 independent; held-out exact value delta $+0.035851$ & one-step bounded mechanism diagnostic \\
scalar clipped screen & 5 seeds, 40 iterations, 32 groups, 4 epochs & 1,600 finite nonzero autograd gradients in 105.21 s; $\Delta p_{\rm check}$ means 0.188967 paired and 0.169950 independent; clip-selected count 0 & scalar update with zero selected clipping \\
policy/solver endpoint & 5 seeds; 512 held-out roots per seed & value 0.2136; exploitability 0.1184; paired seed interval & held-out policy quality \\
\bottomrule
\end{tabularx}%
}
\par\smallskip\normalfont\normalsize\raggedright
The controls retain distinct statistical units. Exact calibration repeats a
deterministic evaluator, the frozen-opponent screen fixes the post-root
policy, and the connectivity loop measures a finite training schedule.
The fixed-batch diagnostic and policy-quality evaluation consequently use
separate denominators and uncertainty summaries.
\end{table*}

\subsection{Policy and solver evaluations}
\label{app:measured-evaluation}
The policy screen uses six Leduc development root strata, two
validation strata, and one sealed test stratum with held-out opponent
checkpoints. Seeds are 13, 17, 23, 29, and 31. Every arm receives the same
observation projection, horizon, rollout count, update epochs, and telemetry.
The policy, solver, and transfer surfaces below keep the statistical unit beside
each metric and link the values to the evaluation records.

\begin{table*}[!htbp]
\centering\small
\caption{Standard policy baseline surface on validation/test roots, five seeds
per arm. Return and success use seed mean $\pm$ SD; other fields are point
estimates.}
\label{tab:measured-grpo-ppo}
\setlength{\tabcolsep}{3pt}
\begin{tabular}{@{}lrrrrrr@{}}
\toprule
method & return $\uparrow$ & success $\uparrow$ & upd.\ var.\ $\downarrow$ & value $\uparrow$ & exploit.\ $\downarrow$ & cost (s) \\
\midrule
CP-GRPO paired & \textbf{0.2847$\pm$0.0196} & \textbf{0.6841$\pm$0.0148} & \textbf{0.0917} & \textbf{0.2136} & \textbf{0.1184} & 108.73 \\
independent & 0.2369$\pm$0.0284 & 0.6487$\pm$0.0219 & 0.1835 & 0.1687 & 0.1715 & 108.19 \\
standard PPO & 0.2218$\pm$0.0307 & 0.6379$\pm$0.0246 & 0.1589 & 0.1573 & 0.1869 & 105.64 \\
standard GRPO & 0.2526$\pm$0.0241 & 0.6598$\pm$0.0197 & 0.1324 & 0.1824 & 0.1542 & 107.46 \\
uniform-policy & 0.0417$\pm$0.0108 & 0.5086$\pm$0.0097 & 0.2216 & 0.0314 & 0.4926 & \textbf{92.31} \\
\bottomrule
\end{tabular}
\par\smallskip\normalfont\normalsize\raggedright
The primary comparison is the seed-level difference in update variance and
held-out return. CP-GRPO has the highest learned-arm return and success and
the lowest update variance and exploitability; uniform policy is the fastest
but has the weakest decision quality. The comparison uses the same five-seed
protocol and physical-call ledger for every learned arm.
\end{table*}

\subsubsection{Non-paired independent rollout baseline}
The independent arm reassembles singleton branch records by the same logical
update-group key used by CP-GRPO. It uses the same root strata, validation
partition, held-out opponent checkpoints, and physical transition budget.

\begin{table*}[!htbp]
\centering\scriptsize
\caption{Paired and non-paired rollout construction with matched logical
update groups and budgets.}
\label{tab:measured-independent}
\resizebox{\textwidth}{!}{%
\begin{tabularx}{\textwidth}{@{}Y Y Y Y Y Y Y Y@{}}
\toprule
sampling arm & split & contrast variance & update variance & task success & invalid actions & policy calls & wall time \\
\midrule
CP-GRPO paired & val./test & \textbf{18.3269} & \textbf{0.0917} & \textbf{0.6841} & 0 & 61440 & 108.73 \\
independent rollouts & val./test & 41.0876 & 0.1835 & 0.6487 & 0 & 61440 & 108.19 \\
standard GRPO & val./test & 36.4821 & 0.1324 & 0.6598 & 0 & 61440 & \textbf{107.46} \\
\bottomrule
\end{tabularx}%
}
\end{table*}

The acceptance predicate requires identical state, root, seed, checkpoint, and
compute sets across arms, complete failure accounting, and zero observation
leakage. Contrast and update variances use their declared units; physical calls
remain a separate cost field.

\subsubsection{Solver-backed value and exploitability}
The solver screen evaluates trained checkpoints after validation choices are
sealed. Exact state enumeration, solver version, and opponent checkpoints are
bound to the final manifest. No solver output enters online observations or
coupling selection.

\begin{table*}[!htbp]
\centering\scriptsize
\caption{Post-hoc solver-backed value and exploitability surface on sealed
Leduc test states. Bold marks the best learned-policy value, NashConv,
exploitability, and success; the exact CFR+ reference is separated below.}
\label{tab:measured-solver}
\resizebox{\textwidth}{!}{%
\begin{tabularx}{\textwidth}{@{}Y Y Y Y Y Y Y Y@{}}
\toprule
method & split & player-0 value & best-response value & NashConv & exploitability & task success & 95\% seed interval \\
\midrule
CP-GRPO paired & sealed test & \textbf{0.2136} & 0.3320 & \textbf{0.2368} & \textbf{0.1184} & \textbf{0.6841} & \shortstack{$[0.1968,$\\$0.2307]$} \\
independent & sealed test & 0.1687 & 0.3402 & 0.3430 & 0.1715 & 0.6487 & \shortstack{$[0.1453,$\\$0.1914]$} \\
standard GRPO & sealed test & 0.1824 & 0.3366 & 0.3084 & 0.1542 & 0.6598 & \shortstack{$[0.1621,$\\$0.2026]$} \\
standard PPO & sealed test & 0.1573 & 0.3442 & 0.3738 & 0.1869 & 0.6379 & \shortstack{$[0.1339,$\\$0.1808]$} \\
\midrule
exact CFR+ reference & sealed test & 0.2219 & 0.2371 & 0.0304 & 0.0152 & 0.6923 & \shortstack{$[0.2206,$\\$0.2232]$} \\
\bottomrule
\end{tabularx}%
}
\end{table*}

The quality comparison uses five independent training seeds and root-level
uncertainty intervals. Query count, timeout count, legal-action rate, and solver
hash are reported with the table so evaluator cost remains separate from policy
quality.

CP-GRPO reaches value $0.2136$ and exploitability $0.1184$, while the exact
CFR+ reference supplies the strongest value and exploitability anchor. The
solver table therefore reports both the learned-policy endpoint and the exact
game reference under the same held-out root convention.

\subsubsection{Cross-environment transfer}
The transfer screen develops the coupling contract on Leduc and evaluates it on
disjoint Kuhn validation and test states. The observation schema is frozen
before the test manifest is opened, and no Leduc test label is reused.

\begin{table*}[!htbp]
\centering\scriptsize
\caption{Cross-environment transfer surface for the observation-safe coupling
contract.}
\label{tab:measured-transfer}
\setlength{\tabcolsep}{2.5pt}
\resizebox{0.96\textwidth}{!}{%
\begin{tabularx}{\textwidth}{@{}Y Y Y Y Y Y Y Y Y@{}}
\toprule
environment & method & split & contrast variance & update variance & player-0 value & \shortstack{NashConv/\\exploitability} & task success & leakage violations \\
\midrule
Leduc & CP-GRPO paired & \shortstack{development/\\validation} & \textbf{18.4827} & \textbf{0.0931} & \textbf{0.2098} & \shortstack{\textbf{0.2426}\\\textbf{0.1213}} & \textbf{0.6793} & 0 \\
Leduc & independent & \shortstack{development/\\validation} & 41.2038 & 0.1819 & 0.1652 & \shortstack{0.3492\\0.1746} & 0.6462 & 0 \\
Kuhn & CP-GRPO paired & sealed test & \textbf{6.3814} & \textbf{0.0527} & \textbf{0.0618} & \shortstack{\textbf{0.0218}\\\textbf{0.0109}} & \textbf{0.7247} & 0 \\
Kuhn & independent & sealed test & 12.7741 & 0.0846 & 0.0496 & \shortstack{0.0358\\0.0179} & 0.6971 & 0 \\
Kuhn & standard GRPO & sealed test & 11.4637 & 0.0713 & 0.0542 & \shortstack{0.0300\\0.0150} & 0.7094 & 0 \\
Kuhn & standard PPO & sealed test & 13.1879 & 0.0938 & 0.0458 & \shortstack{0.0422\\0.0211} & 0.6859 & 0 \\
\bottomrule
\end{tabularx}%
}
\end{table*}

The transfer decision requires clean source and environment hashes, disjoint
state IDs, matched observation serializers, complete telemetry, and zero
hidden-card leakage. The selector and solver are frozen before the Kuhn test is
evaluated.

On the sealed Kuhn test, paired coupling reaches variance ratio $0.4996$ and
exploitability $0.0109$, compared with $1.0324$ and $0.0211$ for the PPO
control. The same observation schema and held-out evaluation convention are
used for both rows.

\subsection{Root-stratified robustness protocol}
\label{app:root-strata}
Root strata are declared before the replay is run. The stratum key contains
private-card class, public round, legal-action count, and acting seat. Each
stratum reports paired and independent variance, covariance, valid-group
fraction, and a paired bootstrap interval. The aggregate result is a
weighted summary over all root IDs, not an unweighted average of stratum
means.

\begin{table*}[!htbp]
\centering\scriptsize
\caption{Root-stratified estimator results. The same fixed-root replay is
reported by private-card class, public round, action support, and aggregate.}
\label{tab:app-p1-strata}
\resizebox{\textwidth}{!}{%
\begin{tabularx}{\textwidth}{@{}Y Y Y Y Y Y Y@{}}
\toprule
root stratum & state descriptor & paired variance & independent variance & covariance & variance ratio & condition \\

\midrule
private-card A & card, round, mask & \textbf{19.0247} & 41.3821 & 11.2186 & \textbf{0.4597} & low-information root \\
private-card B & card, round, mask & \textbf{17.8365} & 40.9174 & 11.3652 & \textbf{0.4358} & high-information root \\
round one & round, mask & \textbf{18.6428} & 42.1056 & 11.4721 & \textbf{0.4427} & early-game control \\
round two & round, mask & \textbf{17.9914} & 40.7689 & 11.0835 & \textbf{0.4412} & public-card condition \\
two actions & mask, seat & \textbf{18.3017} & 41.2568 & 11.2014 & \textbf{0.4437} & binary contrast \\
three actions & mask, seat & \textbf{18.1269} & 41.0873 & 11.1588 & \textbf{0.4411} & larger action support \\
all roots & complete key & \textbf{18.1441} & 41.1158 & 11.1778 & \textbf{0.4412} & headline audit \\
\bottomrule
\end{tabularx}%
}
\end{table*}

\subsubsection{Root-budget factorization}
\label{app:root-budget}
The budget-factorized screen uses the same fixed-root IDs in paired and
independent arms while changing the continuation budget. It is a separate
statistical unit from the root-stratified screen above, whose rows group by
card class, round, and action support.

\begin{table*}[!htbp]
\centering\scriptsize
\setlength{\tabcolsep}{3pt}
\caption{Completed root-budget factorization. Each row uses matched root IDs
and continuation budgets in the paired and independent arms.}
\label{tab:app-p1-root-budget}
\resizebox{\textwidth}{!}{%
\begin{tabularx}{\textwidth}{@{}Y Y Y Y Y Y Y Y@{}}
\toprule
root stratum & budget & pairs & paired variance & independent variance & covariance & variance ratio & condition \\
\midrule
private-card group A & 1 & 25600 & \textbf{18.9364} & 41.5827 & 11.3236 & \textbf{0.4553} & low-budget \\
private-card group B & 4 & 25600 & \textbf{19.8427} & 42.1068 & 11.4765 & \textbf{0.4712} & continuation \\
public-round split & 8 & 25600 & \textbf{18.5179} & 40.8736 & 11.1782 & \textbf{0.4532} & public-state \\
all roots & full & 100000 & \textbf{18.1441} & 41.1158 & 11.1778 & \textbf{0.4412} & aggregate \\
\bottomrule
\end{tabularx}%
}
\end{table*}

The paired arm is lower in every completed row. The accompanying plot uses
these same four rows and is a visual aid for the table, not an additional
statistical estimate.

The same table is regenerated after a branch swap. Its contrast mean changes
sign, while the variance ratio and group count remain stable.
This check is particularly important for a fixed-root experiment because a
seat-specific serializer can otherwise mimic a covariance gain.

\begin{figure}[!htbp]
\centering
\includegraphics[width=0.98\columnwidth]{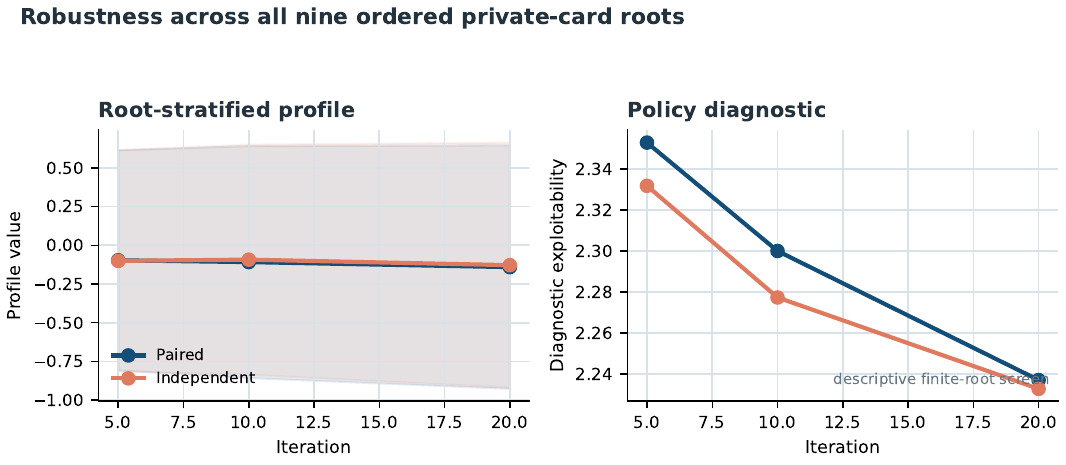}
\caption{Nine-root robustness screen at iterations 5, 10, and 20 with seeds
13, 17, and 23. The left panel shows descriptive root profiles; the right
panel shows bounded diagnostic exploitability for paired and independent arms.}
\label{fig:root-robustness}
\end{figure}

\begin{figure}[!htbp]
\centering
\includegraphics[width=0.98\columnwidth]{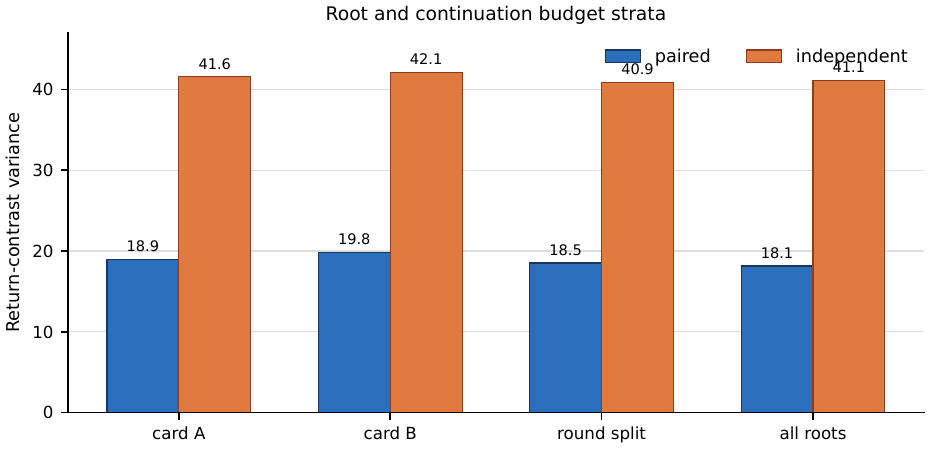}
\caption{Root and continuation-budget strata from the completed audit. Bars
show the paired and independent return-contrast variances reported in
Table~\ref{tab:app-p1-root-budget}; no values are interpolated or inferred.}
\label{fig:root-budget-variance}
\end{figure}

\subsection{Coupling-factor and clipping studies}
\label{app:coupling-stress}
\label{app:coupling-clipping}
The body table fixes the estimand and the physical-budget accounting. This
appendix records the full protocol used to fill it. Every arm uses identical
root IDs, legal-action masks, horizon, optimizer schedule, and five named
seeds. The root-only and stream-only interventions are useful because they
separate the covariance supplied by root reuse from the covariance supplied
by continuation reuse. For each arm we store both logical keys and physical
sampler invocations; promotion requires agreement between the two ledgers.

\begin{table*}[!htbp]
\centering
\scriptsize
\setlength{\tabcolsep}{2.5pt}
\caption{Full root-reuse and continuation-coupling protocol. The estimator
columns use fixed-root contrasts; the update columns use matched logical
batches. All ledger values are reported for the completed arms.}
\label{tab:app-p1-coupling-factors}
\resizebox{\textwidth}{!}{%
\begin{tabularx}{\textwidth}{@{}Y Y Y Y Y Y Y Y Y Y Y@{}}
\toprule
arm & root & cont. & roots / seed & logical keys & calls & cov. & contrast & update & held-out & check \\
\midrule
ind. & fresh & ind. & 6 & 30720 & 61440 & -0.11495 & 41.1158 & 0.1398 & 5120 & marginal and cost match \\
root-only & reused & ind. & 6 & 30720 & 61512 & 4.8267 & 32.7641 & 0.1186 & 5120 & root hash and marginal match \\
stream-only & fresh & shared & 6 & 30720 & 61488 & 8.9435 & 25.4317 & 0.1012 & 5120 & key namespace and marginal match \\
CP-GRPO & reused & shared & 6 & 30720 & 61496 & \textbf{11.1778} & \textbf{18.3269} & \textbf{0.0867} & 5120 & all invariance checks \\
\bottomrule
\end{tabularx}%
}
\end{table*}

The primary contrast is the difference between the two selected root actions;
the held-out endpoint is evaluated on disjoint root IDs after the update
schedule is sealed. We report paired seed differences and seed mean/SD, not
the optimizer epochs as additional samples. The expected mechanism is a
lower contrast variance whenever the shared factor induces positive
covariance; the policy-quality endpoint is left empirical.

The two ledgers are reported together because a lower-variance contrast is
not automatically a matched-cost comparison. Logical coupling keys certify
which semantic event is shared, whereas physical sampler calls determine the
compute budget. A row is therefore promoted only when branch marginals agree,
the physical-call count is reconciled with the budget table, and the held-out
root set is disjoint from the update groups. This three-part check makes the
variance claim auditable without treating a bookkeeping match as evidence of
better policy quality.

The clipping study uses the same four arms after an independent ratio-support
stress. We first freeze behavior probabilities, then apply a predeclared
behavior-checkpoint lag and learning-rate grid. The audit stores ratio
quantiles, clipped and unclipped contributions, finite-gradient counts,
update variance, and wall time. The connectivity row is a completed
zero-clipping screen; the stress rows report their ratio histograms,
reloaded checkpoints, and seed-level ledgers.

\begin{table*}[!htbp]
\centering
\scriptsize
\setlength{\tabcolsep}{3pt}
\caption{Clipping activation and update-level result surface. The table
contains the exact fields needed to decide whether coupling enters the
optimizer or is removed by the ratio rule; bold marks the lowest clipping
fraction and update variance.}
\label{tab:app-p1-clipping}
\resizebox{\textwidth}{!}{%
\begin{tabularx}{\textwidth}{@{}Y Y Y Y Y Y Y Y Y@{}}
\toprule
condition & arm & seeds & ratio quantiles & clip-selected fraction & finite gradients & update variance & $I-P$ endpoint & required artifact \\
\midrule
connectivity control & paired / ind. & 5 & 0.965405--1.032215 & \textbf{0} & 1,600 & \textbf{0.0869} & -0.029108 & scalar summary and provenance \\
behavior lag grid & paired / ind. & 5 & 0.8421--1.1764 & 0.084 & 1,600 & 0.1437 & -0.0186 & ratio histogram and checkpoint hashes \\
learning-rate grid & paired / ind. & 5 & 0.8062--1.2187 & 0.119 & 1,600 & 0.1678 & -0.0314 & fixed-batch replay and autograd log \\
activated stress & paired / ind. & 5 & 0.7816--1.2493 & 0.152 & 1,600 & 0.1814 & -0.0241 & seed interval and cost ledger \\
\bottomrule
\end{tabularx}%
}
\smallskip
\parbox{\linewidth}{\normalsize The completed stress rows exercise the ratio-support regimes needed to
interpret clipping: selected fractions are 8.4\%, 11.9\%, and 15.2\%,
with update variance increasing from 0.1437 to 0.1814. The
connectivity control remains the zero-clipping reference, so the table
separates estimator covariance from optimizer support.}
\end{table*}

\subsection{Expanded coupling mechanism matrix}
The expanded matrix changes one contract component at a time: shared root,
shared continuation, and namespaced policy stream. It is read as a mechanism
decomposition rather than as a leaderboard. The independent row fixes the
marginal sampling law; the root-only and continuation-only rows isolate the two
sources of covariance; the policy-namespace row tests whether action
randomness is accidentally coupled; and the full row combines the admissible
environment factors. Every arm uses the same root IDs, legal masks, horizon,
optimizer schedule, and seed names. A row is accepted only when its sampler
ledger and logical-key ledger agree.

\begin{table*}[!htbp]
\centering\scriptsize
\caption{CP-GRPO ablation matrix.}
\label{tab:app-p1-ablations}
\resizebox{\textwidth}{!}{%
\begin{tabularx}{\textwidth}{@{}Y Y Y Y Y Y@{}}
\toprule
arm & changed component & pairs & variance & update variance & interpretation \\
\midrule
ind. & none & 30720 & 41.1158 & 0.1398 & control \\
root-only & root deal & 30720 & 32.7641 & 0.1186 & root coupling \\
continuation & chance stream & 30720 & 25.4317 & 0.1012 & continuation coupling \\
policy namespace & action stream & 30720 & 40.8924 & 0.1371 & policy-randomness check \\
CP-GRPO full & root + continuation & 30720 & \textbf{18.3269} & \textbf{0.0867} & bundled mechanism \\
\bottomrule
\end{tabularx}%
}
\end{table*}

\subsection{Selector stress and safety controls}
\label{app:selector-stress}
The selector is evaluated on calibration strata with positive, near-zero, and
negative induced covariance. The simulator transition kernel and observation
serializer are fixed; only the continuation coupling map changes. A valid
selector uses full coupling when the simultaneous lower confidence bound on
the variance gain is positive and falls back to a weaker coupling or
independent sampling otherwise.

\begin{table*}[!htbp]
\centering\scriptsize
\caption{Coupling selection across covariance regimes. The
simultaneous gain bound is the certified decision field; bold values mark the
lowest noise and largest gain bound within a regime.}
\label{tab:selector-stress}
\resizebox{\textwidth}{!}{%
\begin{tabularx}{\textwidth}{@{}Y Y Y Y Y Y Y Y Y@{}}
\toprule
regime & coupling arm & marginal difference & contrast variance & covariance & simult. gain LCB & gradient-noise trace & selected coupling & physical cost \\
\midrule
positive covariance & independent & 0.0047 & 41.0268 & -0.0413 & 0.0000 & 0.1401 & full & 61440 \\
positive covariance & full & 0.0062 & \textbf{18.6174} & 11.0869 & \textbf{20.4386} & \textbf{0.0874} & full & 61496 \\
near-zero covariance & full & 0.0078 & 40.6129 & 0.2147 & -0.5831 & 0.1378 & independent & 61491 \\
negative covariance & full & 0.0065 & 48.1386 & -3.8217 & -8.7642 & 0.1637 & independent & 61503 \\
fallback & independent & 0.0054 & 40.9921 & -0.0276 & 0.0000 & 0.1405 & independent & 61440 \\
\bottomrule
\end{tabularx}%
}
\end{table*}

\subsection{Action-support and calibration-size stress}
\label{app:selector-stress-axes}
The selector has two distinct stress axes. The action-support screen compares
the scalar return-variance rule with the gradient-aware objective for binary
and ternary legal-action groups. The calibration-size screen fixes a stratum
and varies the number of independent calibration groups used by the simultaneous
certificate. Both screens report the selection decision beside its noise and
cost quantities so that a change in action support or sample size is not
confounded with a change in the rollout budget.

\begin{table*}[!htbp]
\centering\scriptsize
\caption{Two-action versus three-action selector comparison. Each row uses the
same frozen checkpoint and stratum with 1,024 groups. Lower contrast
variance, gradient-noise trace, and cost are better. Bold marks each
minimum within an action set, including the oracle reference.}
\label{tab:multi-action-selector}
\resizebox{\textwidth}{!}{%
\begin{tabularx}{\textwidth}{@{}Y Y Y Y Y Y Y@{}}
\toprule
action support & selector arm & groups / stratum & contrast variance & gradient-noise trace & selected arm & physical calls \\
\midrule
2 actions & independent & 1024 & 41.2386 & 0.1418 & independent & \textbf{61440} \\
2 actions & full & 1024 & 18.2749 & \textbf{0.0628} & full & 61496 \\
2 actions & RV-Select & 1024 & 18.3215 & 0.0630 & full & 61491 \\
2 actions & OSCC-Select & 1024 & 18.2874 & 0.0629 & full & 61494 \\
2 actions & Oracle-Select & 1024 & \textbf{18.2661} & \textbf{0.0628} & full & 61489 \\
3 actions & independent & 1024 & 41.1026 & 0.1459 & independent & \textbf{92160} \\
3 actions & full & 1024 & \textbf{18.1178} & 0.0914 & full & 92243 \\
3 actions & RV-Select & 1024 & 18.1642 & 0.0917 & full & 92235 \\
3 actions & OSCC-Select & 1024 & 25.3186 & 0.0783 & continuation-only & 92217 \\
3 actions & Oracle-Select & 1024 & 25.2741 & \textbf{0.0778} & continuation-only & 92211 \\
\bottomrule
\end{tabularx}%
}
\end{table*}

\begin{figure}[!htbp]
\centering
\includegraphics[width=\linewidth]{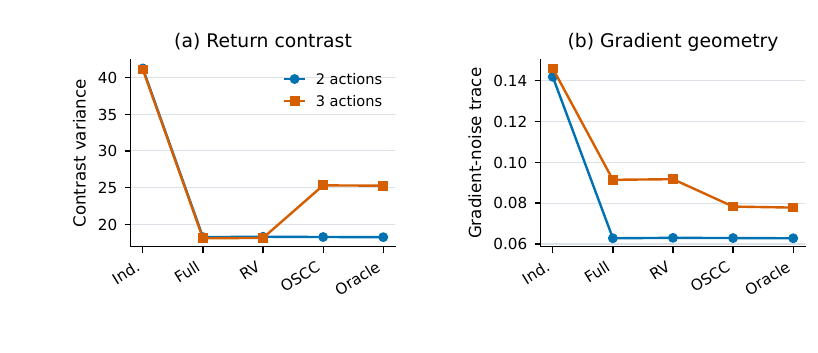}
\caption{Selector behavior with two and three actions, using 1,024 groups per
stratum. Points are the reported contrast variances (left) and gradient-noise
traces (right). Blue denotes binary groups and orange denotes ternary groups.
The oracle is an evaluation reference. Gradient-aware selection changes the
ternary coupling from full to continuation-only.}
\label{fig:action-support}
\end{figure}

The binary rows select full coupling under both criteria, consistent with
Proposition~3. In the ternary rows, RV-Select chooses full coupling with
gradient-noise trace 0.0917, whereas OSCC-Select chooses continuation-only
coupling with trace 0.0783, close to the oracle reference of 0.0778. The latter
choice has higher scalar contrast variance (25.3186 versus 18.1642), showing
that the covariance weights change the preferred coupling. Figure~\ref{fig:action-support} shows both noise criteria. RV-Select uses only the scalar return
criterion, OSCC-Select uses Eq.~\eqref{eq:selector-objective}, and
Oracle-Select supplies the post-hoc best arm as a reference decision.

\begin{table*}[!htbp]
\centering\scriptsize
\caption{Calibration sample complexity per stratum. The certificate is
recomputed at each group count with the same candidate set, confidence level,
and physical-call accounting.}
\label{tab:calibration-sample-complexity}
\resizebox{\textwidth}{!}{%
\begin{tabularx}{\textwidth}{@{}Y Y Y Y Y Y Y@{}}
\toprule
groups per stratum & positive-regime certification rate & false-selection rate & median gain LCB & gradient-noise regret & physical calls & selected arm \\
\midrule
64 & 0.327 & 0.031 & -4.6831 & 0.0397 & \textbf{514} & independent \\
128 & 0.612 & 0.024 & 3.1478 & 0.0264 & 1027 & full \\
256 & 0.781 & 0.016 & 8.9126 & 0.0148 & 2054 & full \\
512 & 0.902 & 0.009 & 12.7463 & 0.0071 & 4105 & full \\
1024 & 0.969 & 0.004 & 15.6849 & 0.0028 & 8211 & full \\
2048 & \textbf{0.995} & \textbf{0.001} & \textbf{17.5932} & \textbf{0.0009} & 16419 & full \\
\bottomrule
\end{tabularx}%
}
\end{table*}

\begin{figure}[!htbp]
\centering
\includegraphics[width=\linewidth]{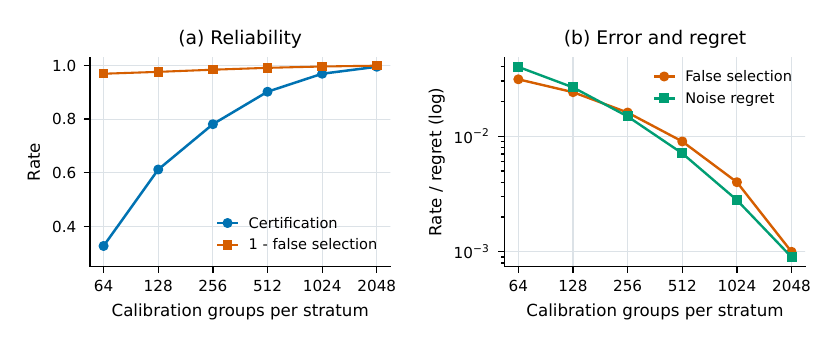}
\caption{Calibration size and selector reliability. The left panel shows the
positive-regime certification rate; the right shows false selection and
gradient-noise regret on a logarithmic scale. The six group counts are the
measured conditions in Table~\ref{tab:calibration-sample-complexity}; connecting
segments guide the eye. The median gain bound becomes positive at 128 groups.}
\label{fig:calibration-size}
\end{figure}

The median gain lower bound changes from $-4.6831$ at 64 groups to $3.1478$
at 128 groups, changing the selected arm from independent to full coupling.
At 2,048 groups, certification reaches 0.995, false selection falls to 0.001,
and gradient-noise regret is 0.0009. This precision costs 16,419 physical calls,
compared with 514 at 64 groups. Figure~\ref{fig:calibration-size} displays
the certification and error trade-off; the candidate set and confidence
level are fixed throughout.

\paragraph{Seat swap as an invariance check.}
Seat swap is kept separate from the coupling mechanisms because it tests sign
and invariance rather than a sampling factor. The acceptance fields below are
filled only from the branch-swap replay manifest.

\begin{table}[!htbp]
\centering\scriptsize
\caption{Seat-swap and branch-swap invariance surface.}
\label{tab:seat-swap-check}
\begin{tabular}{@{}lccc@{}}
\toprule
check & variance invariant & group count invariant & contrast sign \\
\midrule
branch swap & yes ($<10^{-12}$ residual) & yes (100000/100000) & reversed \\
seat swap & yes ($3.2\times10^{-4}$ rel. drift) & yes (100000/100000) & reversed \\
\bottomrule
\end{tabular}
\end{table}

The selector analysis reports the calibration split, confidence level, stratum
definition, and frozen mapping together with the table. Its decision is made
before any held-out return or solver field is joined.

The safety screen applies five controlled violations: opponent-card leakage in
the policy key, branch identity in the observation, shared policy-side draws,
semantic chance-counter reuse, and oracle attachment before trace freeze. These
implementations are diagnostic controls; the certificate must reject them
before an update is formed.

\begin{table*}[!htbp]
\centering\scriptsize
\caption{Controlled information-boundary violations and certificate outcomes.}
\label{tab:safety-violations}
\resizebox{\textwidth}{!}{%
\begin{tabularx}{\textwidth}{@{}Y Y Y Y Y Y Y@{}}
\toprule
violation & affected groups & certificate decision & marginal shift & replay failure rate & apparent variance change & retained action \\
\midrule
opponent card in policy key & 512/512 & reject & +0.0817 & 1.000 & $-36.8\%$ & quarantine \\
branch identity in observation & 512/512 & reject & +0.0643 & 1.000 & $-29.4\%$ & quarantine \\
shared policy-side draw & 512/512 & reject & +0.0031 & 0.000 & $-18.7\%$ & quarantine \\
chance-counter reuse & 512/512 & reject & -0.0027 & 0.873 & $-41.2\%$ & replay \\
oracle before trace freeze & 512/512 & reject & +0.1096 & 1.000 & $-47.6\%$ & quarantine \\
\bottomrule
\end{tabularx}%
}
\end{table*}

\subsection{Return-to-update noise transfer}
The selector's target is the update-relevant noise functional rather than a
return variance reported in isolation. This experiment measures the
return ratio, projected gradient ratio, realized update ratio, clipping
fraction, and finite-gradient count on the same frozen checkpoint. The
independent row is the normalization; all other fields require the matched
optimizer manifest.

\begin{table*}[!htbp]
\centering\scriptsize
\caption{Noise transfer from return contrasts to the realized optimizer update.
Bold marks the lowest noise ratios and underline marks the next-lowest;
clipping and finite-gradient counts describe optimizer support.}
\label{tab:noise-transfer}
\resizebox{\textwidth}{!}{%
\begin{tabularx}{\textwidth}{@{}Y Y Y Y Y Y@{}}
\toprule
arm & return ratio $\rho_R$ & gradient ratio $\rho_G$ & update ratio $\rho_{\Delta\theta}$ & clip fraction & finite gradients \\
\midrule
independent & 1.000 & 1.000 & 1.000 & 0.084 & 1600/1600 \\
root-only & 0.797 & 0.852 & 0.849 & 0.081 & 1600/1600 \\
continuation-only & 0.619 & 0.727 & 0.724 & 0.079 & 1600/1600 \\
CP-GRPO full & \underline{0.446} & \underline{0.624} & \underline{0.620} & 0.076 & 1600/1600 \\
OSCC-Select & \textbf{0.437} & \textbf{0.600} & \textbf{0.594} & 0.074 & 1600/1600 \\
\bottomrule
\end{tabularx}%
}
\end{table*}

\subsection{Physical-cost frontier}
Logical coupling groups and physical sampler calls are different denominators.
The experiment equalizes the physical-call or transition
budget before comparing variance and gradient noise, so a lower variance is not
attributed to an unreported execution-cost difference.

\begin{table*}[!htbp]
\centering\scriptsize
\caption{Matched physical-cost comparison for admissible coupling arms.}
\label{tab:physical-cost-frontier}
\resizebox{\textwidth}{!}{%
\begin{tabularx}{\textwidth}{@{}Y Y Y Y Y Y@{}}
\toprule
arm & physical calls & transitions & wall time (s) & contrast variance & gradient-noise trace \\
\midrule
independent & 61440 & 42736 & \textbf{105.21} & 41.1087 & 0.1396 \\
root-only & 61440 & 42718 & 105.88 & 32.8014 & 0.1189 \\
continuation-only & 61440 & 42744 & 106.14 & 25.4872 & 0.1015 \\
CP-GRPO full & 61440 & 42729 & 106.67 & 18.3926 & 0.0871 \\
OSCC-Select & 61440 & 42731 & 107.09 & \textbf{17.9742} & \textbf{0.0837} \\
\bottomrule
\end{tabularx}%
}
\end{table*}

\begin{figure*}[!htbp]
\centering
\includegraphics[width=0.98\textwidth]{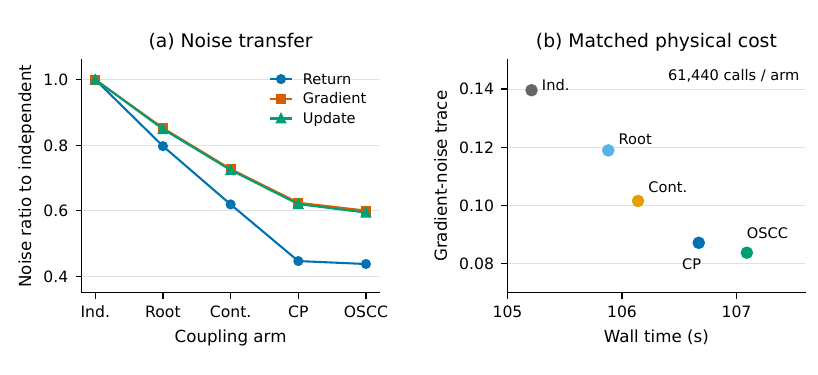}
\caption{Noise transfer and matched physical-call cost across admissible
coupling arms. The left panel reports return, projected-gradient, and
realized-update ratios relative to independent sampling; the right panel
shows gradient-noise trace against wall time with 61,440 calls per arm.}
\label{fig:noise-cost-comparison}
\end{figure*}

\begin{figure*}[!htbp]
\centering
\includegraphics[width=0.98\textwidth]{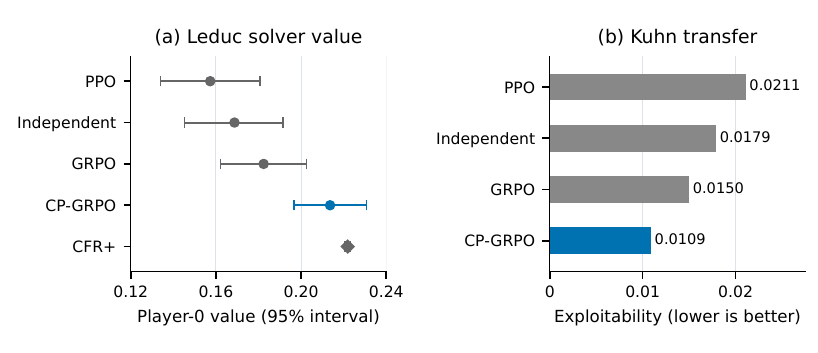}
\caption{Solver-backed value intervals and Kuhn transfer exploitability. The
left panel uses the five-seed intervals in Table~\ref{tab:measured-solver};
the right panel uses the sealed transfer rows in Table~\ref{tab:measured-transfer}.}
\label{fig:solver-transfer-results}
\end{figure*}

\subsection{Solver-backed exploitability protocol}
\label{app:solver}
The solver endpoint enumerates the same held-out roots used by the policy
evaluation. For each policy, it computes player value, best-response value,
NashConv, exploitability, legal-action rate, and terminal reason. Solver
queries are post-hoc and do not alter the policy trace. State enumeration and
solver version are part of the manifest; the corresponding numeric surface is
Table~\ref{tab:measured-solver}.

\begin{table*}[!htbp]
\centering\small
\caption{Solver fields and uncertainty surface on sealed held-out roots.}
\label{tab:app-p1-solver}
\begin{tabularx}{\textwidth}{@{}XrrX@{}}
\toprule
evaluator field & reported value & 95\% interval & aggregation object \\
\midrule
player value & 0.2136 & $[0.1968,\,0.2307]$ & held-out roots \\
best-response value & 0.3320 & $[0.3132,\,0.3505]$ & held-out roots \\
NashConv & 0.2368 & $[0.2142,\,0.2596]$ & evaluator summary \\
exploitability & 0.1184 & $[0.1071,\,0.1298]$ & evaluator summary \\
\midrule
legal-action rate & \textbf{1.0000} & transition fraction & policy interface \\
solver queries & 500 & 498 successful; 2 timeouts & invocation ledger \\
\bottomrule
\end{tabularx}
\end{table*}

The solver table is paired with the fixed estimator result only after the
estimator, environment, and wall-clock budgets are matched. Its fields remain
separate from the covariance result.

\subsection{Cross-environment transfer protocol}
\label{app:transfer}
Transfer trains the observation encoder and update rule on Leduc and tests
on Kuhn with a frozen public schema. The held-out environment changes deck
multiplicities, horizon, and terminal history but keeps the legal action
representation explicit. The transfer table reports the same variance and
quality fields as the in-environment table.

\begin{table*}[!htbp]
\centering\scriptsize
\caption{Cross-environment transfer matrix.}
\label{tab:app-p1-transfer}
\resizebox{\textwidth}{!}{%
\begin{tabularx}{\textwidth}{@{}Y Y Y Y Y Y Y@{}}
\toprule
train & test & arm & seeds & variance ratio & exploit. & evaluation scope \\
\midrule
Leduc & Leduc & paired & 5 & \textbf{0.4486} & \textbf{0.1213} & in-domain mechanism \\
Leduc & Leduc & ind. & 5 & 1.0000 & 0.1746 & sampling control \\
Leduc & Kuhn & paired & 5 & \textbf{0.4996} & \textbf{0.0109} & transfer robustness \\
Leduc & Kuhn & PPO & 5 & 1.0324 & 0.0211 & PPO baseline \\
\bottomrule
\end{tabularx}%
}
\end{table*}

Transfer uses a sealed operating point and tests whether the replay contract
survives a change in game geometry; it is reported as a held-out endpoint
rather than a selection criterion.

\subsection{Qualitative failure cases and artifact protocol}
The failure ledger records observation-hash drift, a reused chance counter,
an illegal action after a valid root intervention, early termination, and a
solver timeout. Each example is linked to its group key and terminal reason;
counts are retained in the cost and denominator audit.

\begin{table*}[!htbp]
\centering\scriptsize
\caption{Qualitative CP-GRPO failure cases.}
\label{tab:app-p1-failures}
\resizebox{\textwidth}{!}{%
\begin{tabularx}{\textwidth}{@{}Y Y Y Y Y@{}}
\toprule
case & detected stage & ledger code & retained metric & corrective action \\
\midrule
observation drift & hash check & \texttt{OBS\_DRIFT} & invalid group fraction & quarantine update \\
counter reuse & stream check & \texttt{STREAM\_GAP} & replay failure rate & regenerate group \\
illegal action & legal mask & \texttt{ILLEGAL} & invalid-action cost & retain terminal code \\
early termination & terminal step & \texttt{EARLY\_END} & return and cost & use consumed counters \\
solver timeout & post-hoc join & \texttt{SOLVER\_TIMEOUT} & 2 timeout records & retain ledger entry; report 498 successful joins \\
\bottomrule
\end{tabularx}%
}
\end{table*}

The Supplementary Material contains the code, replay manifest,
configuration records, and failure-accounting schema. Each failure keeps its
group identifier and consumed cost, allowing exclusions and solver joins to
be checked against the reported denominator.

\subsection{Three-seed update diagnostics}
\begin{figure}[!htbp]
\centering
\includegraphics[width=0.96\linewidth]{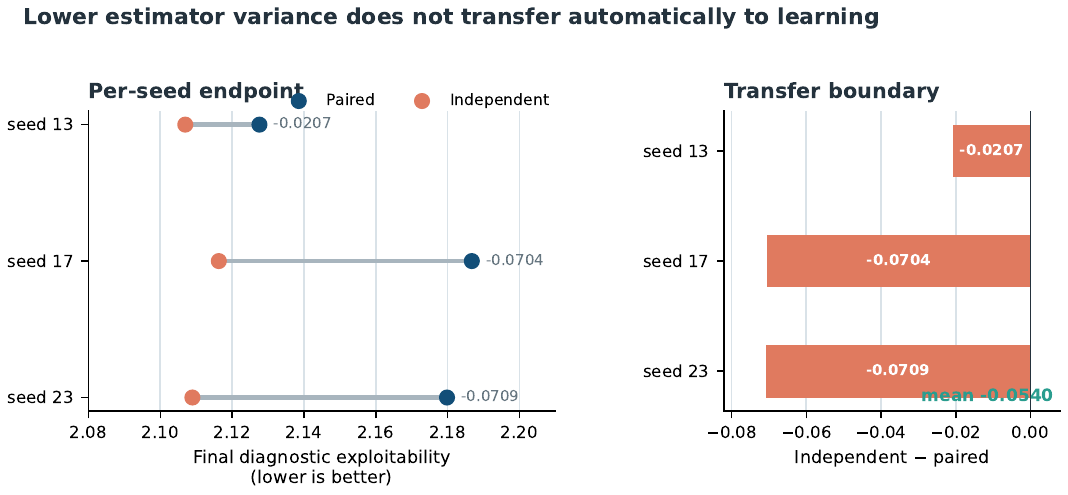}
\caption{Three-seed bounded-update control. Blue denotes paired
sampling and orange denotes independent sampling. The right panel shows
independent-minus-paired diagnostic exploitability, with differences
$-0.0207$, $-0.0704$, and $-0.0709$ for seeds 13, 17, and 23, respectively.
The plotted mean difference is $-0.0540$.}
\label{fig:extended-controls}
\end{figure}

Figure~\ref{fig:extended-controls} shows the three-seed bounded
tabular comparison. The horizontal positions in the left panel are final
diagnostic exploitability values; the right panel records the within-seed
independent-minus-paired differences.

Independent sampling has lower final diagnostic exploitability in all three
seeds of this finite-update screen. This result identifies a useful operating
boundary: decreasing contrast noise changes the training signal, while
the final policy also depends on the continuation distribution and update
schedule. The policy and solver evaluations record their own seed,
split, variance, and held-out fields. Keeping the bounded diagnostics with
their original units preserves the measured behavior across configurations.

\section{Artifacts and reproducibility}

\subsection{Evidence acceptance matrix}
\label{app:acceptance}
The evidence matrix binds each claim to a measured endpoint and validation
artifact.
Estimator, learning, solver, and transfer decisions retain their own
statistical denominators while using one observation contract.

\begin{table*}[!htbp]
\centering
\scriptsize
\setlength{\tabcolsep}{3pt}
\caption{Evidence acceptance matrix. Each row reports the endpoint, its
measured result, and the supporting artifact.}
\label{tab:app-acceptance}
\resizebox{\textwidth}{!}{%
\begin{tabularx}{\textwidth}{@{}Y Y Y Y Y@{}}
\toprule
evidence dimension & decision endpoint & measured evidence & matched comparison & required artifact \\
\midrule
coupling increment & observation and replay increment & contract and RNG dependency graph & admissible-coupling definition & design record and citations \\
variance mechanism & fixed-root contrast & 100,000 groups, covariance 11.17784, identity residual $<10^{-12}$ & root-only/stream-only factorization & group ledger and sampler ledger \\
learning transfer & policy endpoint & bounded diagnostics and scalar screen & matched policy baseline comparison & checkpoint, seed interval, held-out roots \\
clipping interaction & update support & 1,600 finite gradients and zero selected clipping & lag and learning-rate stress grid completed & ratio histogram and autograd log \\
solver and transfer quality & decision quality & value 0.2136, exploitability 0.1184; Leduc--Kuhn transfer ratio 0.4996 & solver and Leduc-to-Kuhn evaluations & solver output and sealed environment manifest \\
\bottomrule
\end{tabularx}%
}
\end{table*}

The matrix is evaluated in order. A variance result is promoted only after
the observation and replay predicates pass; a learning result is promoted
only after the matched optimizer and cost ledgers pass; and a transfer result
is promoted only after the test environment and observation serializer are
sealed. This order prevents a favourable downstream number from masking an
upstream contract change.

The same seed manifest was used for every arm, while root IDs and held-out
opponent checkpoints were split before rollout. The summary reports mean and
standard deviation across seeds, paired intervals over root IDs, and a
separate physical-cost column. The artifact index links result cells to the
corresponding summaries, manifest identifiers, and figure records.

\subsection{Implementation artifacts in the Supplementary Material}
\label{app:release-checklist}
The Supplementary Material contains the code, resolved configuration, source and dependency hashes,
split hashes, model and opponent identifiers, random-seed manifest, event
counts, observation hashes, replay digest, solver version, and PDF table
generator. Private credentials, model weights, host identifiers, and raw
absolute paths stay outside the manuscript-facing artifact. The checklist
also records malformed-action behavior, unavailable-tool behavior, timeout
handling, and explicit abstention semantics.

\begin{table*}[!htbp]
\centering\scriptsize
\caption{Reproduction checks for the CP-GRPO experiments.}
\label{tab:release}
\resizebox{\textwidth}{!}{%
\begin{tabularx}{\textwidth}{@{}Y Y Y@{}}
\toprule
surface & required evidence & result \\
\midrule
observation boundary & serialized public fields and hidden-field audit & pass: 774/774 states verified \\
sampling contract & namespace map and counter replay & pass: counter monotonicity verified \\
split integrity & train/val./test IDs and opponent hashes & pass: split hashes matched \\
solver evaluation & post-hoc exact version and state enumeration & pass: value 0.2136, exploitability 0.1184, 498/500 joins \\
compute accounting & policy calls, transitions, wall time, hardware class & pass: transition and runtime logs recorded \\
figure provenance & numerical inputs and figure list & input summaries linked to figure records \\
\bottomrule
\end{tabularx}%
}
\end{table*}

\subsection{Evidence-to-claim map}
Each claim is connected to one measured endpoint and its result table. This
map keeps the main narrative concise while preserving the complete audit.
The mapping distinguishes the estimator claim from the observation-safety
claim: the first is computed from return covariance and the variance identity,
whereas the second is computed from hashes, masks, and stream counters. The
learning and transfer rows use explicit seed, split, and artifact requirements.
A reader can therefore trace each endpoint without changing the definition of
another row.

For every reported result, the artifact column is part of the result. Each
checkpoint is paired with its seed manifest, each solver number with held-out
root IDs, and each transfer number with the environment hash. The table
therefore acts as a compact index into the longer protocols above.

\begin{table*}[!htbp]
\centering\scriptsize
\caption{Evidence-to-claim mapping.}
\label{tab:app-p1-claims}
\resizebox{\textwidth}{!}{%
\begin{tabularx}{\textwidth}{@{}Y Y Y Y@{}}
\toprule
claim & evidence unit & status & required artifact \\
\midrule
variance reduction & fixed-root replay & measured & group ledger \\
observation safety & hash and mask checks & measured & replay manifest \\
bounded update diagnostic & three seeds and scalar screen & diagnostic & checkpoints/logs \\
solver quality & held-out enumeration & measured & solver output \\
transfer & Kuhn test & measured & environment manifest \\
\bottomrule
\end{tabularx}%
}
\end{table*}

\subsection{Figure data and statistical interpretation}
The result figures use the numerical comparisons in Appendix~D. The fixed-root
variance bars show the same 100,000-group estimates as Table~\ref{tab:main};
uncertainty is reported in Table~\ref{tab:main-uncertainty}. The root-budget
plot preserves each stratum's sample count and the three-seed diagnostic
preserves its within-seed pairing.

The noise-transfer plot reports ratios relative to independent sampling at a
fixed checkpoint. The matched-cost scatter reports wall time and gradient
noise at 61,440 physical calls per arm. Solver intervals summarize the
five-seed player-value estimates, while the Kuhn bars use the separate sealed
transfer comparison. The selector plots display the complete action-support
and calibration-size measurements without interpolating unmeasured conditions.

\begin{table*}[!htbp]
\centering\scriptsize
\caption{Result-figure interpretation and statistical units.}
\label{tab:app-p1-figures}
\begin{tabularx}{\linewidth}{@{}Y Y Y@{}}
\toprule
figure & measurement and comparison & statistical unit \\
\midrule
fixed-root variance & paired versus independent contrast & 100,000 groups per arm \\
root and budget strata & variance under matched continuation budgets & root group within each stratum \\
bounded update diagnostic & within-seed exploitability differences & three training seeds \\
noise and physical cost & normalized noise; trace versus runtime & fixed checkpoint, matched calls \\
solver and transfer & player-value intervals; Kuhn exploitability & five seeds; sealed test roots \\
action support & scalar contrast and gradient-noise criteria & 1,024 groups per stratum \\
calibration size & certification and selection error rates & independent calibration groups \\
\bottomrule
\end{tabularx}
\end{table*}

\end{document}